\documentclass[10pt,journal]{IEEEtran}

\usepackage{caption}
\usepackage{amsmath,graphicx}
\usepackage[export]{adjustbox}
\usepackage{subfig}
\usepackage{siunitx}
\usepackage{cancel}

\usepackage{cite}

\ifCLASSINFOpdf
\else
\fi

\usepackage{tikz}
\usepackage{pgfplots}

\usetikzlibrary{shapes,arrows,arrows.meta,positioning}
\pgfplotsset{compat=newest}
\pgfplotsset{plot coordinates/math parser=false}

\usepackage{amsmath}
\usepackage{amssymb}
\usepackage{bm}
\DeclareMathOperator*{\argmin}{arg\,min}  %
\DeclareMathOperator*{\argmax}{arg\,max}
\DeclareMathOperator*{\mymax}{max}

\usepackage{microtype}

\usepackage{booktabs}

\usepackage{url}

\usepackage{siunitx}
\usepackage{acronym}

\newcommand{\vect}[1]{\bm{#1}}
\colorlet{blue}{black}
\newcommand{\rev}[1]{\textcolor{blue}{#1}}

\begin{document}
\title{Deep Learning Methods for Vessel Trajectory Prediction based on Recurrent Neural Networks}

\author{Samuele~Capobianco,
        Leonardo~M.~Millefiori,~\IEEEmembership{Member,~IEEE,}
        Nicola~Forti,
        Paolo~Braca,~\IEEEmembership{Senior~Member,~IEEE}
        and~Peter Willett,~\IEEEmembership{Fellow,~IEEE}%
\thanks{\copyright 2021 IEEE.  Personal use of this material is permitted.  Permission from IEEE must be obtained for all other uses, in any current or future media, including reprinting/republishing this material for advertising or promotional purposes, creating new collective works, for resale or redistribution to servers or lists, or reuse of any copyrighted component of this work in other works.}%
\thanks{S. Capobianco was with the NATO Science and Technology Organization (STO) Centre for Maritime Research and Experimentation (CMRE), 19126 La Spezia, Italy. He resides in Monsummano Terme (PT), Italy. E-mail: research@samuelecapobianco.com.}%
\thanks{L. M. Millefiori, N. Forti and P. Braca are with the NATO Science and Technology Organization (STO) Centre for Maritime Research and Experimentation (CMRE), 19126 La Spezia, Italy. E-mail: \{name.surname\}@cmre.nato.int. }%
\thanks{P. Willett is with the Department of Electrical and Computer Engineering, University of Connecticut, Storrs, 06269. E-mail: peter.willett@uconn.edu.}%
\thanks{This work is supported by the NATO Allied Command Transformation (ACT) via the project ``Data Knowledge Operational Effectiveness'' (DKOE).}%
}

\maketitle

\newacro{ACT}{NATO Allied Command Transformation}
\newacro{AIS}{Automatic Identification System}
\newacro{CMRE}{Centre for Maritime Research and Experimentation}
\newacro{COG}{Course Over Ground}
\newacro{DKOE}{Data Knowledge Operational Effectiveness}
\newacro{GPS}{Global Positioning System}
\newacro{GT}{gross tonnage}
\newacro{IMO}{International Maritime Organization}
\newacro{MSA}{Maritime Situational Awareness}
\newacro{MMSI}{Maritime Mobility Service Identity}
\newacro{PoL}{Pattern Of Life}
\newacroplural{PoL}[PoL]{Patterns Of Life}
\newacro{SOG}{Speed Over Ground}
\newacro{SOLAS}{International Convention for the Safety of Life at Sea}
\newacro{UTM}{Universal Transverse Mercator}
\newacro{SaR}{Search and Rescue}
\newacro{SAR}{Synthetic Aperture Radar}
\newacro{MSE}{Mean Squared Error}
\newacro{DMA}{Danish Maritime Authority}
\newacro{nmi}{nautical miles}
\newacro{NCV}{Nearly Constant Velocity}
\newacro{OU}{Ornstein-Uhlenbeck}
\newacro{NN}{Neural Network}
\newacroplural{NN}[NNs]{Neural Networks}
\newacro{RNN}{Recurrent Neural Network}
\newacroplural{RNN}[RNNs]{Recurrent Neural Networks}
\newacro{LSTM}{Long Short-Term Memory}
\newacro{kn}{knots}
\newacro{EMODnet}{European Marine Observation and Data Network}
\newacro{AOI}{area of interest}

\begin{abstract}
Data-driven methods open up unprecedented possibilities for maritime surveillance %
using Automatic Identification System (AIS) data. 
In this work, 
we explore deep learning 
strategies 
using historical AIS observations
to address the %
problem of 
predicting future vessel trajectories
with a prediction horizon %
of several hours.
We propose novel sequence-to-sequence vessel trajectory prediction models 
based on encoder-decoder recurrent neural networks (RNNs)
that are trained on historical trajectory data
to predict future trajectory 
samples given previous observations.
The proposed architecture 
combines Long Short-Term Memory (LSTM) RNNs 
for sequence modeling to encode the observed data and generate future predictions with
different intermediate aggregation layers to capture space-time dependencies in sequential data.
Experimental results on vessel trajectories from 
an AIS dataset made freely available by the Danish Maritime Authority show
the effectiveness of 
deep-learning methods 
for %
trajectory 
prediction 
based on sequence-to-sequence neural networks,
which achieve better performance than baseline approaches 
based on linear regression or \rev{on the Multi-Layer Perceptron (MLP) architecture}. %
The comparative evaluation of results 
shows: %
i) the superiority of %
attention pooling over static pooling for the specific application,
and ii)
the 
remarkable
performance improvement
that can be obtained
with labeled trajectories,
i.e., when predictions are conditioned on a 
low-level context representation encoded from 
the sequence of 
past observations, 
as well as
on additional inputs
(e.g., port of departure or arrival) 
about the vessel's high-level intention 
which 
may be available from AIS. %
\end{abstract}


\IEEEpeerreviewmaketitle

\section{Introduction}
\label{sec:intro}

\begin{figure}[!t]
    \centering
    \includegraphics[trim=130 10 130 10,clip,width=1\columnwidth]{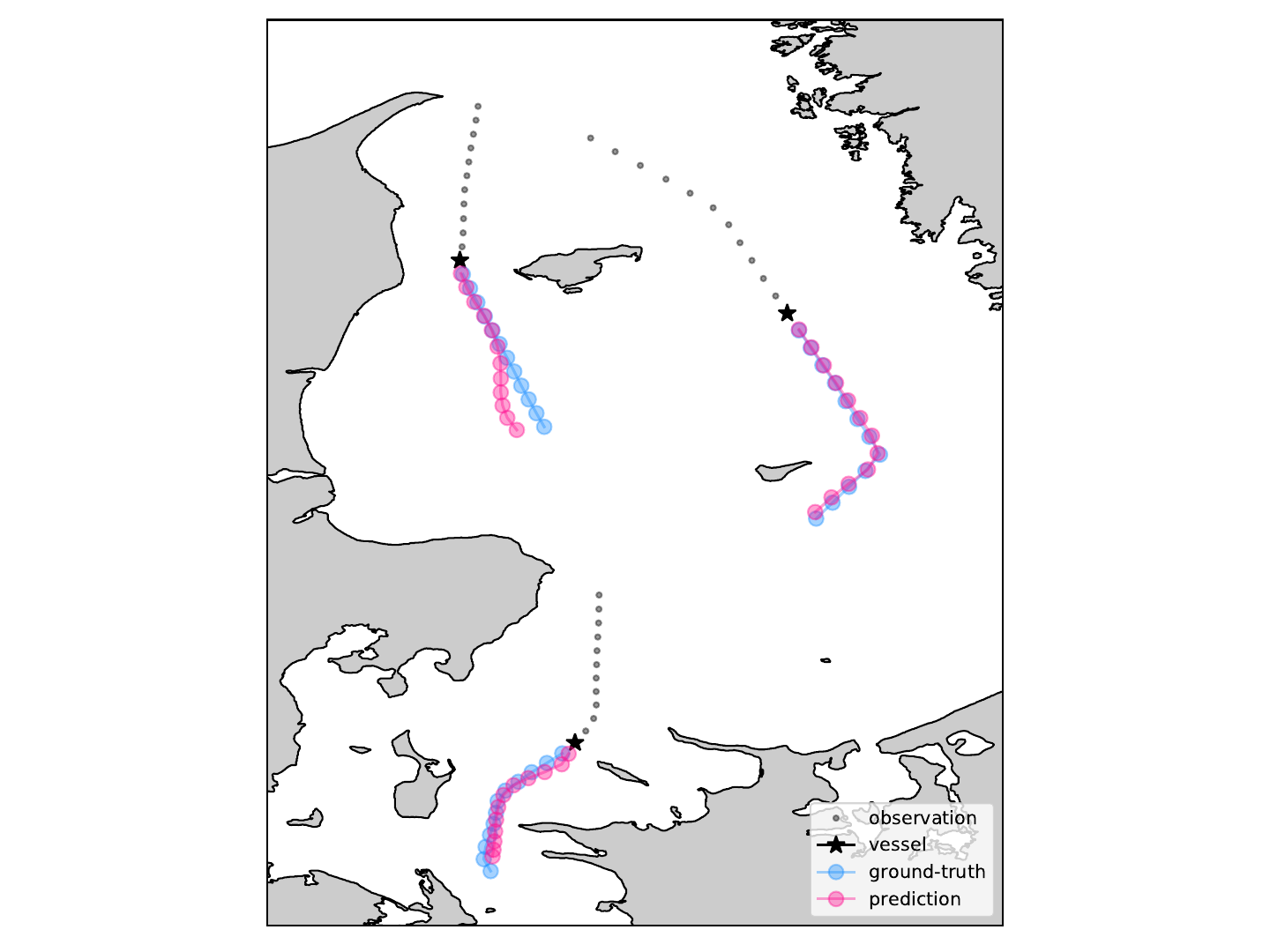}
    \caption{Selected trajectories from the dataset used in this paper. The last observed vessel position is marked with a black star; the predicted and actual trajectory are depicted in pink and blue, respectively. \rev{Contains data from the Danish Maritime Authority that is used in accordance with the conditions for the use of Danish public data~\cite{DMA}.}}
    \label{fig:predictions_intro}
\end{figure}

Technology adoption of intelligent integrated maritime safety systems in the near future
is expected to reduce the number of vessel accidents and improve safety in the maritime environment, especially as autonomous shipping is being proactively explored in the maritime industry.
A key 
capability of such systems is to predict the future positions of vessels, even 
over extended prediction horizons, and also %
anticipate ships' future behavior to enhance awareness of imminent hazards. Such capability is not only needed for future intelligent integrated maritime safety systems, where autonomous and non-autonomous, commercial and recreational, large- and small-sized ship traffic will have to coexist, but it is also essential in other applications, e.g., early emergency response\footnote{For example, in search and rescue operations and man-overboard situations, where vessels closest to the accident might have not updated their position in hours.} or prevention of piracy attacks\footnote{In areas at high risk of piracy, ship masters have the option of turning off safety systems, such as the AIS, if they believe 
the safety or security of the ship could be compromised. In these situations, 
the position of a vessel can be established using 
predictive techniques, even if it is temporarily unobservable.}. Furthermore, advanced predictive techniques enable the association of %
ship observations from heterogeneous, non-synchronized sensors and support the capability of intercepting vessels of interest from satellite platforms. Satellite acquisitions need to be scheduled in advance, and, if the vessel of interest is in motion, a prediction capability is required to define the acquisition area around the probable location of the vessel when the satellite will pass over it.

Maritime surveillance systems 
are increasingly relying on the vast amount of data 
made available by terrestrial and satellite networks of \ac{AIS}
stations designed to receive radio messages broadcast by ships with their position, identity, %
and other voyage-related information. %

Systems similar to the \ac{AIS} are also acutely needed in the
automotive field %
to monitor road traffic in support of autonomous vehicle technology.
The research community has recently devoted significant attention to the problem of behavior prediction for autonomous driving applications; a recent and comprehensive review of the state of the art is presented in~\cite{Mozaffari2020}.
However, while vehicles %
have a %
very constrained motion, governed by road lanes, driving rules, and environmental conditions~\cite{Mozaffari2020},
in the maritime domain vessel trajectories are %
constrained by sea lanes only in specific areas where traffic separation schemes are enforced. Even when it comes to voyage planning, the actual trajectory of a vessel might differ significantly from the planned one, due to sea current, weather and
traffic conditions along the way. %
All these factors, and the fact that the \ac{AIS} still has non-negligible limitations in coverage, latency and reliability,  %
make vessel trajectory prediction a challenging task, %
quite 
distinctive from the automotive application.

At the same time, apart from an ubiquitous monitoring capability, the availability of %
maritime traffic data in such large volumes also
creates the opportunity to systematically extract 
useful knowledge 
to improve the security and safety of maritime operations and enhance %
the overall 
situational awareness.
The use of massive historical AIS data for vessel trajectory prediction remains, however, an open problem, and a challenging one.
This is not only because of 
\ac{AIS} technological limitations, such as the highly irregular time sampling, poor data quality and integrity, but also because the variety of behaviors exhibited by ships (depending on, e.g., their size, navigational status, traffic regulations, etc.).%

To deal with this complexity and address the vessel prediction problem, in this paper we propose a fully data-driven approach, as opposed to a model-based one, able to implement a prediction capability that is effective even in complex and elaborated traffic scenarios.

\subsection{Related work}
Among the model-based prediction tools, the easiest and maybe most common %
is probably the linear \ac{NCV} model~\cite{Rong2003}.
The \ac{NCV} model's strengths are the ability 
to perform short-term predictions 
of straight-line trajectories,
and its inherent robustness with respect to the quality of input data.
However, the \ac{NCV} model is not appropriate for medium- and long-term forecasting, as it tends to
overestimate the actual uncertainty of the prediction, as the time horizon increases.
An only slightly more complex %
model for ship prediction, based on the \ac{OU} stochastic process, has been recently proposed and shown to be particularly well-suited for non-maneuvering motion and
long prediction time windows~\cite{Millefiori2016,Millefiori2015,Uney2019}. 
The \ac{OU} model has also been adopted for maritime anomaly detection~\cite{icassp19} and, in combination to data-driven methods, to develop unsupervised procedures that automatically extract knowledge about maritime traffic patterns~\cite{Millefiori2016-2,Coscia2018,oceans2019}.

Additional works approached the \ac{AIS}-based vessel trajectory prediction problem from a more data-driven perspective, 
exploring adaptive kernel density estimation~\cite{Ristic08},
nearest-neighbor search methods~\cite{Hexeberg17}, nonlinear filtering~\cite{Mazzarella15},
Gaussian Processes~\cite{Rong2019},
and machine learning techniques~\cite{Zissis2017}.
A comprehensive survey on the recent efforts of vessel trajectory prediction using AIS data can be found in~\cite{survey2018} and~\cite{survey2020}.

More recently, advances in deep learning and the combined availability of %
massive volumes of \ac{AIS} data 
are paving the way to enhance vessel trajectory prediction and maritime surveillance. \rev{The main motivation is that neural architectures have the capability to learn complex motion patterns directly from the data, making it possible to predict future vessel kinematic states even in complex traffic scenarios where conventional statistical techniques would struggle to achieve satisfactory performance.}   
Recent examples of deep learning techniques applied to vessel trajectory prediction
include~\cite{Nguyen2018,Gao2018,Yu2020,Murray2020,Zhou2020,Nguyen2018b,Forti2020}.
In~\cite{Nguyen2018}, a variational \ac{RNN} architecture (named \textit{GeoTrackNet}) is proposed to extract knowledge on the hidden behavior of ships from \ac{AIS} data streams
for track reconstruction and anomaly detection. 
The prediction of ship behavior is addressed in~\cite{Gao2018},
by directly using bidirectional \ac{LSTM} \acp{RNN} %
for online forecasting, and in~\cite{Yu2020}, where recurrent neural models are adopted to process sequential AIS data with the final goal of associating incoming messages with existing tracks.
Moreover, in~\cite{Murray2020} the authors present an auto-encoder approach to predict entire vessel trajectories (with prediction horizons up to $30$ minutes), instead of sequential states, based on previous clustering and classification steps, which are required for proper trajectory extraction.
\rev{In the target tracking literature, a recent work~\cite{Jung2020} proposes a neural network architecture based on \ac{LSTM} with uncertainty modeling to incorporate non-Markovian dynamic models in the prediction step of a standard Kalman filter for target tracking.}
Some other works %
divided a surveillance area into spatial grids and applied deep-learning tools to predict vessel inflow and outflow~\cite{Zhou2020}, or maritime trajectories with estimated destination and arrival time~\cite{Nguyen2018b}. 
Using a spatial grid-based approach, in~\cite{Nguyen2018b} the transition of vessel states is defined between cells on the grid, rather than between time intervals of \ac{AIS} observations. This permits assignment of %
a unique code to each cell and represent a vessel trajectory
as a text sequence, {which can be used} 
to train a sequence-to-sequence model similar to those %
{widely used for} neural machine translation 
problems~\cite{Cho2014,Sutskever2014}.

Indeed, neural encoder-decoder models 
have recently become a standard
approach for sequence-to-sequence tasks such as machine translation~\cite{Cho2014,Sutskever2014}
and speech recognition~\cite{Chiu18}.
Such models first use a \ac{RNN} to encode the input sequence as a set of vector representations,
which are then decoded by a second \ac{RNN} to generate the output sequence step-by-step, conditioned on the encodings.

The first work addressing
vessel trajectory prediction from AIS data
through
sequence-to-sequence models based on the LSTM encoder-decoder architecture
is documented in~\cite{Forti2020}, where future trajectory states of a vessel are generated given a sequence of past \ac{AIS} observations.

The remainder of the paper is organized as follows. \rev{In Section~\ref{sec:contribution}, we describe the specific contribution of this work.} In Section~\ref{sec:buildingdata}, we present a method to prepare %
an \ac{AIS} dataset %
to train the proposed neural networks for vessel trajectory prediction. %
In Section~\ref{sec:method}, we formalize the problem and formulate the sequence-to-sequence learning approach to vessel trajectory prediction. Section~\ref{sec:encdec} describes the encoder-decoder neural architecture used to address the prediction task. Experimental results using the proposed dataset are presented and discussed in Section~\ref{sec:experiments}. Finally, \rev{we discuss the limitations and possible future extensions of the proposed approach in Section~\ref{sec:limitations}} and conclude the paper in Section~\ref{sec:conclusion}.

\section{\rev{Contribution}}
\label{sec:contribution}

Building upon\cite{Forti2020}, 
we propose a vessel trajectory learning and prediction framework to generate future trajectory samples
of a vessel given a sequence of past \ac{AIS} observations. 
The idea is to build an end-to-end prediction system able to learn the mapping between sequences of past and future vessel states. 
Inspired by the recent success of deep-learning approaches for sequence-to-sequence applications, 
the proposed method is based on the \ac{LSTM} encoder-decoder architecture, which has emerged as an effective and scalable model for sequence-to-sequence learning. 

\rev{This work goes beyond~\cite{Forti2020} in several regards. We employ a different dataset of vessel trajectories, which is not only larger, but also significantly more complex. Moreover, we investigate here the use of several intermediate aggregation layers (including, but not limited to, the attention mechanism) and the exploitation of high-level intention information (i.e., ship destination). Finally, the architecture described here features several other improvements with respect to~\cite{Forti2020} as, for example, the use of a bidirectional encoder recurrent network and a custom initialization of the decoder network's hidden state~\eqref{eq:decoder_init}. Finally, the model hyper-parameters are chosen with an optimization procedure that aims at finding the most suitable configuration.}

We adopt \ac{LSTM} RNNs~\cite{Hochreiter1997} for sequence modeling to encode context representation from past data and generate future predictions.
The ability to learn from data with long-range temporal dependencies makes \ac{LSTM} RNNs a natural choice for trajectory prediction tasks, given the time lag existing between the input samples and the outputs to be predicted.
In addition, we study how encoder and decoder networks can be effectively connected through intermediate aggregation layers capturing space-time dependencies in sequential data. 
Different from standard vessel prediction systems, our solution is designed to deal with maneuvering motion and long prediction time windows in complex dynamics of maritime traffic. %

A conventional approach for predicting the behavior of a vessel is usually based exclusively on the information contained in the track history of vessel states (e.g., position, velocity) or summarized in the current state without any information related to the ship's intended journey. 
This work explores how it is also possible to extend the input information of the deep-learning model by considering, in addition to a sequence of past vessel states, the prior ship's %
long-term intention (e.g., departure and arrival port) 
which 
is commonly available 
in the received AIS messages. 
A similar 
extension can be found
in~\cite{Ristic08}, where the authors showed that
better results in terms of prediction accuracy
can be obtained by labelling a vessel trajectory based on its long-term intention.
Using the labeled trajectories, we cope with the uncertainty in %
the problem,
realizing a system able to forecast future trajectories at long time horizons in complex maritime traffic environments.

To summarize, we propose a comprehensive deep-learning approach 
for the task of vessel trajectory prediction based on:
i) extracting vessel motion patterns from large volumes of \ac{AIS} data, 
and 
ii) training \acp{RNN} 
to sequentially predict a %
vessel trajectory, given a sequence of \ac{AIS} observations. 
In particular,
our work focuses on:
\subsubsection{Extracting vessel routes from AIS data}%
Historical \ac{AIS} data usually {comes in the form of} a collection of messages for a given time window and area of interest. However, the set of available \ac{AIS} data does not contain information about the specific motion pattern each message is associated with. We propose a simple and effective solution to {aggregate trajectories in sets, so that the components of each set all belong to the same motion pattern.}%
\subsubsection{Establishing a novel deep-learning framework} 
We propose a novel architecture for sequence-to-sequence vessel trajectory prediction based on encoder-decoder LSTM \acp{RNN} with different aggregation layers able to capture complex space-time dependencies in sequential \ac{AIS} data. The proposed framework can also exploit non-kinematic information (e.g., destination) to further improve the prediction performance.
\subsubsection{Comparing different deep learning methods}
We explore different deep-learning methods to {perform} vessel trajectory prediction and compare the performance of sequence-to-sequence models based on \acp{RNN} against feed-forward architectures. 
Moreover, we show how it is possible to exploit external pattern knowledge (e.g., destination information) by %
using additional features in the predictive model. 
We show the effect of such non-kinematic features on trajectory prediction for different architectures. In the following, we will refer to the models that use external features as \textit{labeled} models; conversely, with \textit{unlabeled} models we will refer to models that use only kinematic features.

\section{Maritime traffic patterns}
\label{sec:buildingdata}

\begin{figure}
\centering
  \subfloat[EMODnet]{
  \includegraphics[trim=220 18 220 18,clip,width=0.45\columnwidth]{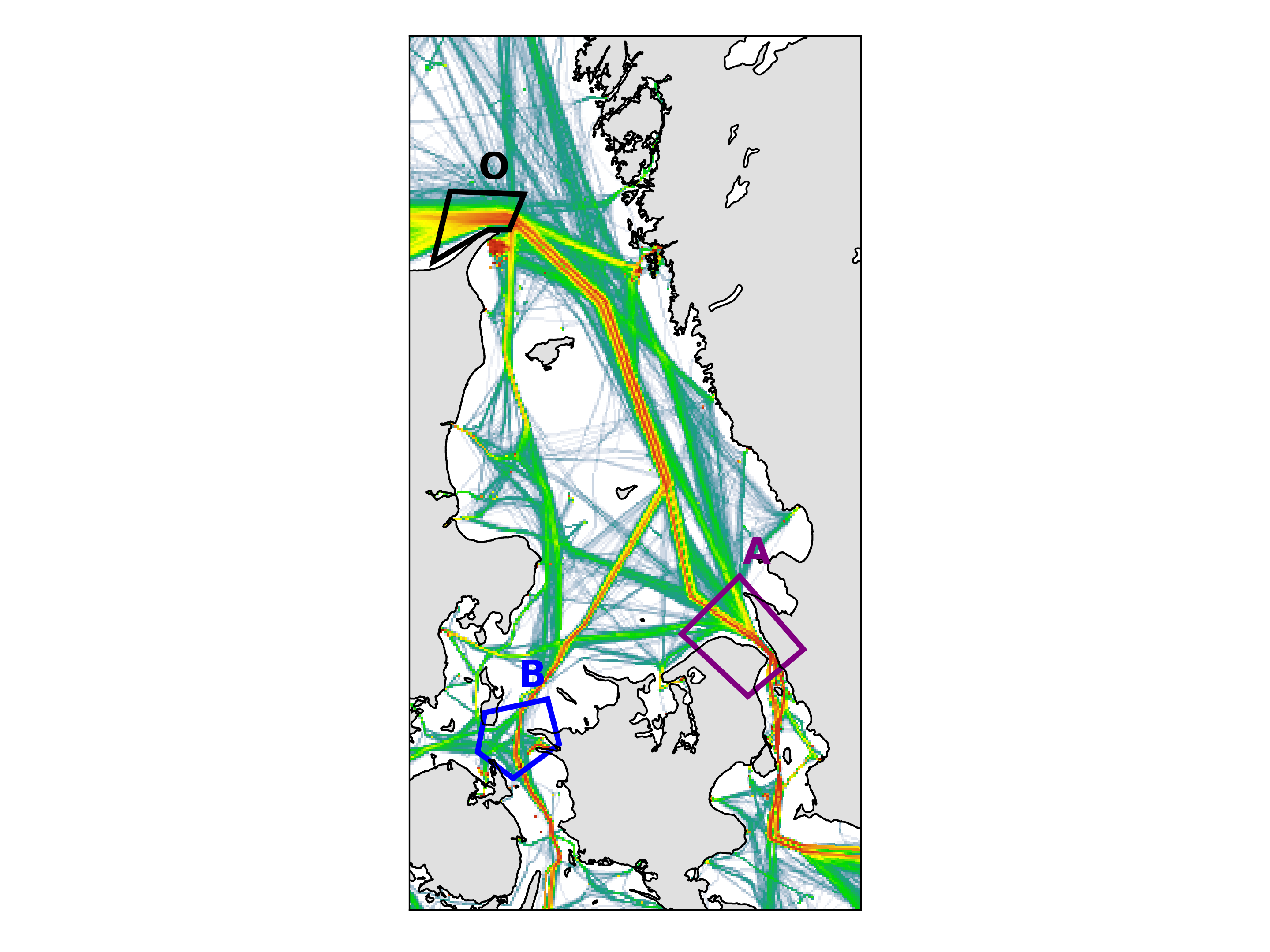}
  \label{fig:emodnet}
} 
  \subfloat[Extracted data]{
  \includegraphics[trim=220 18 220 18,clip,width=0.45\columnwidth]{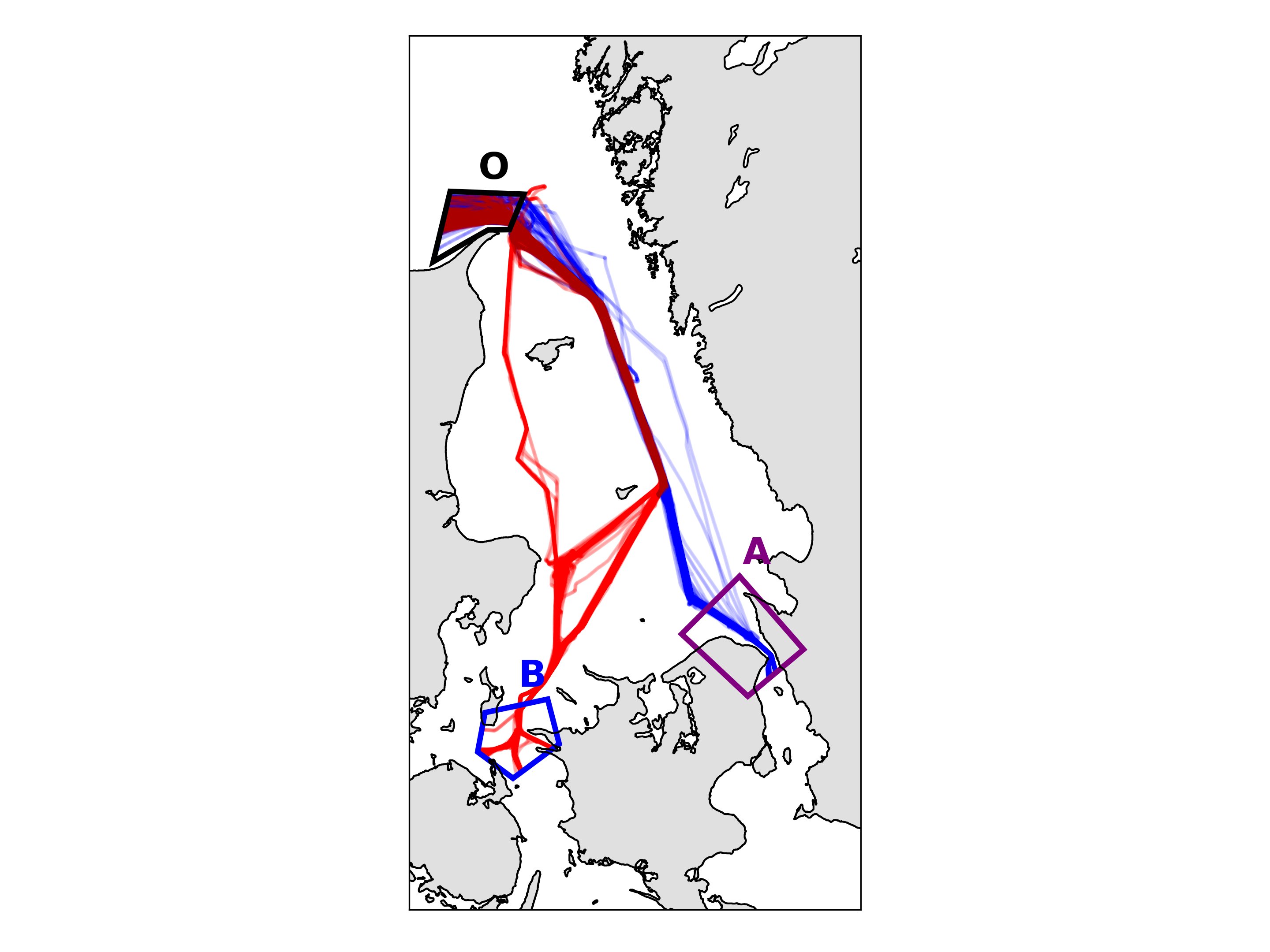}
    \label{fig:tanker_density}
}   
\caption{Vessel density maps in Denmark waters. \protect\subref{fig:emodnet}~Vessel density map from EMODnet with the defined PGAs. \protect\subref{fig:tanker_density}~\ac{AIS} messages belonging to two predefined vessel motion patterns---$(O, A)$ in blue, $(O, B)$ in red---selected using the method described in
Section~\ref{sec:buildingdata}. \rev{Contains data from the Danish Maritime Authority that is used in accordance with the conditions for the use of Danish public data~\cite{DMA}.}}
\label{fig:density_maps}
\end{figure}

The analysis of spatial traffic patterns is of interest for several applications in many domains, from transportation and surveillance, to wireless mobile networks, where mobility modeling is needed to properly address the complexity of the agents' motion within the system. %
In the maritime domain, historical ship mobility data, e.g., from the AIS, %
reveals, for instance, %
commercial maritime routes and how shipping lanes %
articulate in national and international %
waters. 

An intuitive tool to visualize the spatial distribution of \ac{AIS} data, and hence ship traffic, are density maps. %
The simplest way to create a density map is building
a grid over an \ac{AOI} and compute, for each grid cell, the number of \ac{AIS} messages that ships broadcast from it in a given time frame.
\rev{More advanced methods to create density maps, such as
those freely available on the ``Human Activities'' portal of the \ac{EMODnet}, a network of organizations supported by the EU's integrated maritime policy, use a first stage that aggregates AIS positions in ship trajectories. Then, they rely on the number of intersections between ship trajectories and the grid to estimate the traffic density~\cite{EMODNET_detail}.} 
An example of \ac{AIS} density map is illustrated in Fig.~\ref{fig:emodnet}, made with data taken from the \ac{EMODnet} Human Activities portal~\cite{EMODNET_HA}, and shows the vessel traffic density map in Danish waters in March 2019.%

Figure~\ref{fig:emodnet} conveys an important message: unsurprisingly, vessel traffic is inherently regular; in other words, vessels are not \textit{randomly} distributed in the water space, but rather tend to follow seemingly regular paths, driven by, e.g., traffic regulations and minimization of fuel consumption. Moreover, typically vessels plan their voyage from the starting point to the destination by identifying intermediate positions, namely waypoints, where changes of course will occur; once decided on their navigational plan, vessels then follow these predetermined courses from waypoint to waypoint until they reach their final destination. The aim of our effort is to capitalize on this regularity to learn vessel trajectory patterns in order to compute future vessels' behavior ahead of time, even in complex maritime environments.

Let us define a vessel trajectory as the ordered temporal sequence of positions of a vessel; in practice, the construction of vessel trajectories amounts to aggregating and ordering in time \ac{AIS} positions by the reported \ac{MMSI} number.
For the scope of this work, we are not interested in building a dataset of \textit{generic} vessel trajectories, but one of trajectories sharing a common pattern, for instance a \textit{journey}, i.e., the origin and destination of the voyage. 
To this aim, a simple and effective way to select, among all the trajectories in a dataset, only those satisfying such a requirement is to define a set of %
polygonal geographical areas (PGAs) and select only the trajectories that intersect two PGAs in a given order. In this way, we can isolate sets of trajectories with homogeneous (in space) origin and destination.

Figure~\ref{fig:tanker_density} illustrates an example of a dataset created using this approach, where, starting from a historical dataset of \ac{AIS} messages~\cite{DMA} freely available from the Danish Maritime Authority (DMA), %
we selected only the vessel trajectories belonging to two motion patterns: $(O,A)$, in blue, and $(O,B)$, in red. In Section~\ref{sec:experiments} we will describe how this dataset can be used to train deep-learning models for the task of vessel trajectory prediction.
Note that the one above is not the only possible approach to obtain sets of homogeneous trajectories; %
one could use other information to build a set of \textit{homogeneous} trajectories, for instance the ports of departure and arrival, or could opt for a grid-based approach. %

At this stage it is also possible to assign each trajectory with a pattern description that represents the specific motion pattern the trajectory belongs to. %
In the following, we will use as pattern descriptor the destination PGA, but in a practical application may be the vessel's destination (e.g., as declared in the AIS), the port of departure, or the vessel category.

\section{Trajectory prediction}
\label{sec:method}
In this section, we formally introduce the problem of vessel trajectory prediction  
and a probabilistic formulation of the sequence-to-sequence learning approach used to address the prediction task.
We present a data-driven approach to find an approximate solution 
based on recurrent networks for sequence modeling in order to encode information from past data and generate future predictions. 
\subsection{Problem definition}
\label{sec:sec2_1}

A dataset of
$N$ space-time trajectories 
can be represented by a set of temporally-ordered sequences of tuples $\mathcal{C}=\{\mathcal{S}^i,\mathcal{T}^i,\Psi^i\}_{i=1}^{N}$. Each %
tuple in the sequence
is formed by concatenating 
a sequence $\mathcal{S}^i$ of states, i.e., a trajectory
\begin{equation}\label{eq:traj}
\mathcal{S}^i = \{ \mathbf{s}^i_k; \, k = 1,\dots,T_i  \},
\end{equation} 
a list $\mathcal{T}^i = \{ t_1,\dots,t_{T_{i}} \}$, $t_1 < \dots < t_{T_{i}}$ of time points, and
a categorical feature $\Psi^i \in \{\psi_1, \dots , \psi_P\}$  expressing the class label of the specific motion 
pattern to which the %
trajectory $i$ belongs. 
The categorical feature is an optional input of %
the predictive model, depending on whether this %
information is available (labeled data) or not (unlabeled data).
We can represent it with %
a $P$-way categorical feature, where  
$P$ is the cardinality of the set of all possible motion patterns.
A common choice to represent categorical features is %
one-hot encoding\cite{Hancock2020}, 
by which categorical variables are converted into a 
$1$-of-$P$ binary vector $\bm{\psi}^i \in 
\{0,1\}^{1 \times P}$. 
However, more sophisticated techniques are also available~\cite{piech2015deep}.
Note that 
$T_i$ in~\eqref{eq:traj} is the length (number of samples) of trajectory $i$, 
and each sequence element $\mathbf{s}^i_k \triangleq \mathbf{s}^i(t_k) \in \mathbb{R}^d$ 
represents the state %
on trajectory $\mathbf{\mathcal{S}}^i$ at time $t_k$.
The %
state usually consists of kinematic real-valued features of dimension $d$, including positional and velocity information. %
In our experiments, we populated %
$\mathbf{s}_k$ with %
geographical coordinate pairs (longitude, latitude) of the vessel at time $t_k$, but other choices are possible.
Based on the 
class label $\mathcal{P}$,
the resulting collection $\mathcal{C} \triangleq \cup_{j=1}^{P} \mathcal{C}_j$ of $N = |\mathcal{C}|$ historical trajectories can be defined as the union of all $P$ sets of motion patterns $\mathcal{C}_j$,
each containing $N_j$ training vessel paths
such that
$N = \sum_{j=1}^P N_j$.

Given a training dataset $\mathcal{C}$ of historical %
trajectories %
from different motion patterns,
the problem of 
pattern learning and
trajectory prediction is then to 
learn, from this training set, the underlying spatio-temporal mapping
that enables to predict %
future kinematic states based on previous observations, with a prediction horizon generally in the order of hours. 

In the following %
we will show how we can formulate the %
trajectory prediction problem 
as a supervised learning process 
from sequential data
by following a sequence-to-sequence modeling approach
to directly generate an output sequence of future states given an input sequence of %
observations.
To this end, a data segmentation procedure is first required in order to obtain batches of input/output sequences.

\subsection{Data segmentation}\label{se:segmentation}

\begin{figure}
    \centering
    \includegraphics[trim=80 10 80 10,clip,width=1\columnwidth]{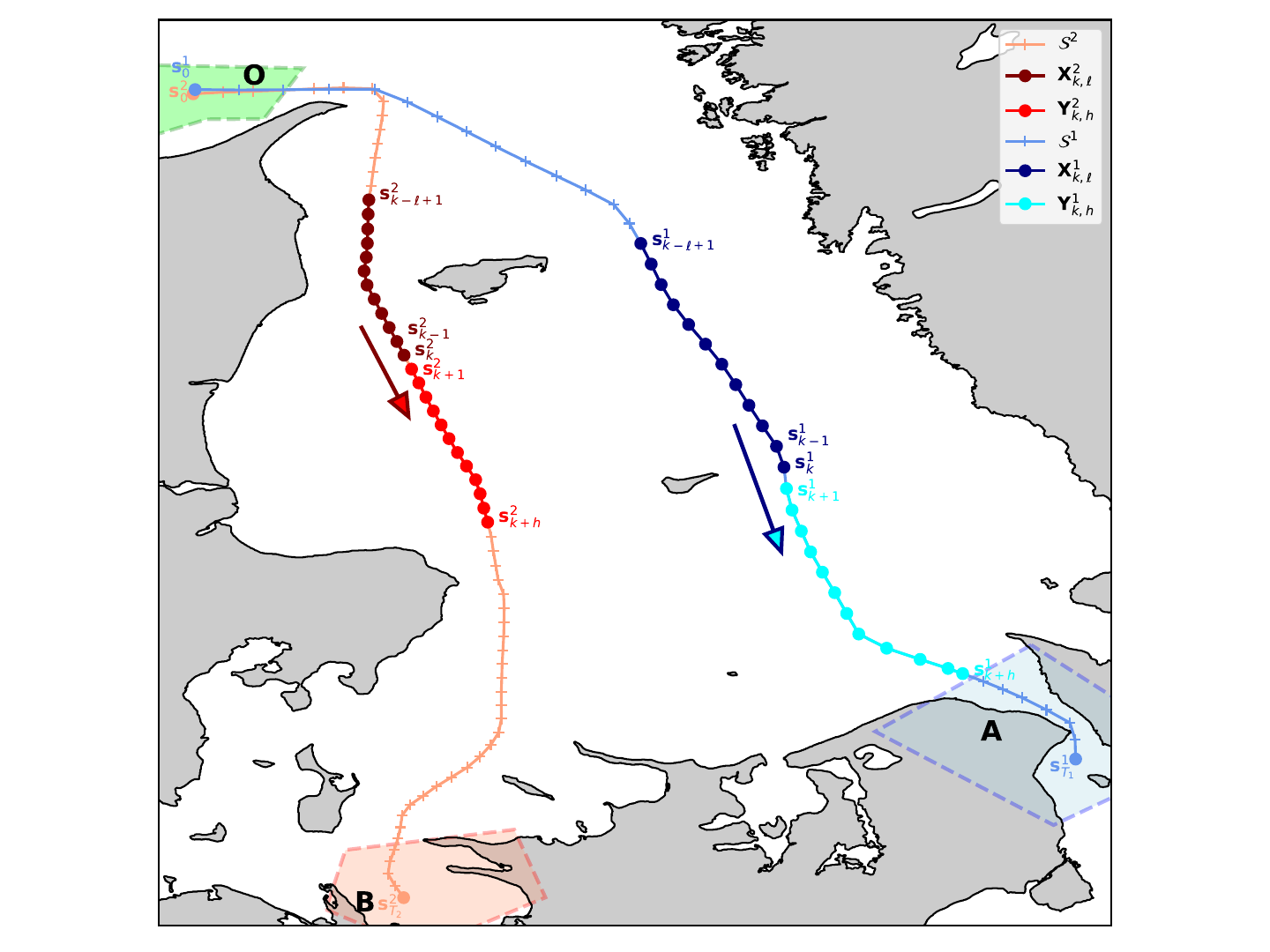}
    \caption{Two examples of observed (input) and future (target) sequences starting at 
    time step $k$ 
    for a prediction task
    of vessel trajectories $\mathcal{S}^1$ and $\mathcal{S}^2$. For each sample, the full trajectory is illustrated with plus markers representing the input to the windowing procedure, which, for each sample in the trajectory, produces an input/output fixed-length sequence pair, depicted in the figure with darker and lighter colors, respectively. \rev{Contains data from the Danish Maritime Authority that is used in accordance with the conditions for the use of Danish public data~\cite{DMA}.}}  
    \label{fig:timeline}
\end{figure}

\rev{The original collection $\mathcal{C}$ of historical trajectories contains sequences with variable length of irregularly-sampled 
vessel 
kinematic states $\mathcal{S}^i$ in~\eqref{eq:traj}.
The kinematic states are irregularly-sampled due to the fact that AIS messages are received at irregular time intervals. }
The first step is therefore to
obtain regularly-sampled trajectories using a fixed sampling time $\Delta$; in other words, we
interpolate the original trajectories at times $t_k=k \, \Delta$, $k=1, \dots, T_i$. 
The  second step is %
to use a fixed-size sliding window (windowing) approach %
to %
reduce the data to arbitrary fixed-length input and output sequences;
this allows performing trajectory prediction given %
a fixed-length sequence of past observations, and translates our problem to a fixed-length input/output mapping learning one.
In this way, the segmentation procedure re-structures the available temporal sequences in $\mathcal{C}$
into a new dataset %
of input/output sequences that can be %
directly used for supervised learning. 
Otherwise stated, segmentation %
is the process of %
splitting whole trajectories (i.e., multivariate time series) %
into smaller, fixed-length sequences. %
The sliding window method is a simple and widely employed segmentation procedure for time series~\cite{Zaroug20},
where previous time steps (input variables) are used  to predict the next time steps (output variables).
Here, the sliding window procedure
produces
an input sequence %
of $\ell$ samples, an output sequence %
of $h$ samples
and has a sliding size of one step. 
Figure~\ref{fig:timeline} depicts %
the sliding window procedure applied to two trajectories at a single time instant. %

In summary, %
the windowing procedure rearranges the interpolated version of $\mathcal{C}$ into a new data representation 
$\mathcal{D}=\{\mathcal{X}^i,\mathcal{Y}^i,\Psi^i\}_{i=1}^{N}$ 
by extracting from each trajectory $i$: 
a set 
$\mathcal{X}^i = \{ \mathbf{X}_{k,\ell}^i \}_{k=\ell}^{T_i-h}$
of $n_i=T_i-(\ell+h)+1$ \rev{sequences} %
such that 
$\mathbf{X}_{k,\ell}^i \triangleq 
\{ \mathbf{x}_{\tau}^i \}_{\tau=k-\ell+1}^{k} =
\{ \mathbf{s}_{\tau}^i \}_{\tau=k-\ell+1}^{k} \subseteq \mathcal{S}^i$
is the input sequence of $\ell \geq 1$ observed states up to time $k$;
a set 
$\mathcal{Y}^i = \{ \mathbf{Y}_{k,h}^i \}_{k=\ell}^{T_i-h}$
such that 
$\mathbf{Y}_{k,h}^i \triangleq 
\{ \mathbf{y}_{\tau}^i \}_{\tau=k+1}^{k+h}=
\{ \mathbf{s}_{\tau}^i \}_{\tau=k+1}^{k+h} \subseteq \mathcal{S}^i$ 
is the output sequence of at time $k$ of $h \geq 1$ future states;
a categorical variable
$\Psi^i$ expressing the class label of trajectory $i$.
Note that, as usually done in sequence-learning problems, we denote the observed and future states with different symbols (i.e., $\mathbf{x}$ and $\mathbf{y}$) even if they belong to the same state space,
in order to better distinguish the input from the output states.
After applying the %
windowing procedure %
to all the trajectories, the new dataset 
$\mathcal{D}$ for sequence-to-sequence trajectory prediction
contains
$N\times n$ tuples, with $n=\sum_{i=1}^N n_i$,
of input sequences $\mathbf{X}_{k,\ell}^i \in \mathbb{R}^{d \times \ell}$, with pattern descriptors $\bm{\psi}^i \in 
\{0,1\}^{1 \times P}$
and output sequences $\mathbf{Y}_{k,h}^i \in \mathbb{R}^{d \times h}$.

\subsection{Sequence-to-sequence modeling approach}\label{se:problem_setup}

We propose a sequence-to-sequence deep-learning approach 
to the problem of %
trajectory prediction introduced in Section~\ref{sec:sec2_1}. 
A sequence-to-sequence model aims to map a fixed-length input sequence with a fixed-length output sequence, where the length of the input and output may differ.
From a probabilistic perspective, the objective is to learn the 
following predictive distribution at time step $k$
\begin{equation}\label{eq:predictive_distr}
p(\mathbf{y}_{k+1}, \dots, \mathbf{y}_{k+h} | \mathbf{x}_{k-\ell+1}, \dots, \mathbf{x}_k, \bm{\psi} ),
\end{equation} 
which represents
the probability of 
a target sequence 
$\mathbf{Y}_{k,h}$ 
of $h$ future states of the vessel
given the
input sequence $\mathbf{X}_{k,\ell}$ 
of $\ell$ observed states, and the possibly available 
journey descriptor feature $\bm{\psi}$.
Once the predictive distribution is learned from the available training data, the neural network architecture can be used to directly sample a target sequence given an input sequence of observed states.

The supervised learning task
can be recast as a sequence regression problem~\cite{Bishop96},
which aims at 
training a neural network model 
$F_{\ell,h}$ 
to predict, given an input sequence $\mathbf{X}_{k,\ell}$ of length $\ell$
and the 
possibly available pattern descriptor 
$\bm{\psi}$,
the output sequence 
$\hat{\mathbf{Y}}_{k,h}$ of length $h$
that maximizes the conditional probability \eqref{eq:predictive_distr}, i.e., 
\begin{equation}\label{eq:pred}
\hat{\mathbf{Y}}_{k,h} = F_{\ell,h} ( \mathbf{X}_{k,\ell}, \bm{\psi} ) = \argmax_\mathbf{\mathbf{Y}} p( \mathbf{Y}_{k,h} | \mathbf{X}_{k,\ell}, \bm{\psi} ).
\end{equation} 
The parameterized function $F_{\ell,h} (\mathbf{X}_{k,\ell}, \bm{\psi}; \bm{\theta} )$ can be directly trained from the given 
dataset 
$\mathcal{D}$
in order to find the set of parameters $\hat{\bm{\theta}}$ that best approximate the mapping function in \eqref{eq:pred} 
over all the input-output training samples, i.e.,
that minimize 
some task-dependent error measure
$\mathcal{L}$:
\begin{equation}\label{eq:lossmin}
  \hat{\bm{\theta}}=\argmin_{\bm{\theta}}\frac{1}{N}\sum_{i=1}^N\mathcal{L}\Big(F_{\ell,h} (\mathbf{X}_{k,\ell}^i, \bm{\psi}^i;\bm{\theta}),\mathbf{Y}_{k,h}^i\Big),
\end{equation}
where $N$ is the number of training samples, %
$\mathbf{X}_{k,\ell}^i$, $\mathbf{Y}_{k,h}^i$, and $\bm{\psi}^i$ are, respectively, the input sequence, target sequence, and pattern descriptor of the $i$-th sample.
$\mathcal{L}$ is the loss function used to measure the prediction error 
which is minimized %
to train the neural network model.

In the following we will see 
a neural network architecture
designed 
to approximate \eqref{eq:pred}, which is 
capable of 
generating, in a recursive way, a sequence of future trajectories
after having captured  %
the underlying motion patterns from historical data. %
Moreover, we will see that the neural architecture can use the ground-truth label $\bm{\psi}$ (e.g., destination information) to extend the context information used to perform predictions. In the case that the label $\bm{\psi}$ is not available, the neural architecture will perform unlabeled predictions based only on the past observed vessel states.

Next, we will briefly review the basic principles of LSTM RNNs, as they are an important element for sequence-to-sequence modeling.
For readers already familiar with this concept, 
we recommend moving on to Section \ref{sec:encdec}.

\subsection{Recurrent networks for sequence modeling}\label{sec:rnns}
RNNs are 
natural extensions of feedforward networks with cyclical connections to facilitate pattern recognition in sequential data.
This structure enables RNNs 
to effectively model temporal dynamic behavior
by summarizing 
inputs into internal hidden states,
used as a memory mechanism to 
capture dependencies
throughout a temporal
sequence.
Such a property makes RNNs ideal %
for sequence modelling.
RNNs have achieved outstanding results in
speech recognition~\cite{Graves2013}, machine translation~\cite{Sutskever2014,Cho2014,bahdanau2015}, and image captioning~\cite{Xu2015}. 

For simplicity, %
let us denote by 
$\mathbf{X}_{\ell} = \{\mathbf{x}_{t}\}_{t=1}^{\ell}$
a generic input sequence of length $\ell$
where the time index $k$ has been dropped, and $t$ now denotes the element position in the sequence such that $\mathbf{x}_{t} = \mathbf{x}_{k-t+1}$.
Then, a generic RNN
sequentially reads each element 
$\mathbf{x}_{t}$
of the input sequence 
and updates its internal hidden state $\mathbf{h}_{t} \in \mathbb{R}^q$ 
according to:
\begin{equation}\label{eq:RNN}
    \mathbf{h}_{t} = g (\mathbf{x}_{t},\mathbf{h}_{t-1}; \bm{\theta}),
\end{equation}
where $g$ is a nonlinear activation function with $\bm{\theta}$ learnable parameters.
RNNs are usually trained with additional networks, 
such as output layers,
that accept as input %
the hidden state $\mathbf{h}_{t}$ in~\eqref{eq:RNN} to make predictions sequentially.

While in theory RNNs are simple and powerful models for sequence modelling, in practice %
they are challenging to train via gradient-based methods, 
due to the difficulty of learning long-range temporal dependencies~\cite{Bengio1994}.
The two major reasons %
behind this behavior are the problems of exploding and vanishing gradients when backpropagating errors across many time steps, meaning gradients that either increase or decrease exponentially fast, respectively, making the optimization task either diverge or extremely slow.  

LSTM networks~\cite{Hochreiter1997} are a special kind of RNN architecture capable of learning
long-term dependencies; they have been successfully applied to language modeling and other applications to overcome the vanishing gradient problem~\cite{Bengio1994}. 
The LSTM architecture consists of a set of memory blocks, each containing 
a cell state and three gates (sigmoidal units):
an input gate to control how the input can alter the cell state, an output gate to set what part of the cell state to output, and a forget gate~\cite{Felix2000} to decide how much memory to keep. 
The input sequence 
$\mathbf{X}_{\ell}$ 
is passed through the
LSTM  network to compute sequentially the hidden vector sequence 
$\mathbf{H}_{\ell} \triangleq \{\mathbf{h}_t\}_{t=1}^{\ell}$. 
In particular, the memory cell takes the input vector at the current time step $\mathbf{x}_{t}$ and the hidden state at the previous time step $\mathbf{h}_{t-1}$
to update the internal hidden state $\mathbf{h}_{t}$ by using the following equations:
\begin{IEEEeqnarray}{rCl}
    \mathbf{i}_{t} &=& \operatorname{\sigma}(\mathbf{U}_{i} \mathbf{x}_{t} + \mathbf{W}_{i} \mathbf{h}_{t-1} + \mathbf{b}_i) \nonumber\\
    \mathbf{f}_{t} &=&  \operatorname{\sigma}(\mathbf{U}_{f} \mathbf{x}_{t} + \mathbf{W}_{f} \mathbf{h}_{t-1} + \mathbf{b}_f) \nonumber\\
    \mathbf{o}_{t} &=&  \operatorname{\sigma}(\mathbf{U}_{o}\mathbf{x}_{t}  + \mathbf{W}_{o} \mathbf{h}_{t-1}  + \mathbf{b}_o) \nonumber\\
    \Tilde{\mathbf{c}}_{t} &=& \operatorname{tanh}(\mathbf{U}_{c} \mathbf{x}_{t} + \mathbf{W}_{c} \mathbf{h}_{t-1} + \mathbf{b}_c) \nonumber\\
    \mathbf{c}_{t} &=& \mathbf{f}_{t}\odot \mathbf{c}_{t-1} + \mathbf{i}_{t}\odot\Tilde{\mathbf{c}}_{t} \nonumber\\
    \mathbf{h}_{t} &=& \mathbf{o}_{t}\odot \operatorname{tanh}(\mathbf{c}_{t}),\label{eq:LSTM}
\end{IEEEeqnarray}
where $\odot$ denotes the element-wise product,
$ \operatorname{\sigma}$ is the  sigmoid function,
$\operatorname{tanh}$ is the hyperbolic tangent function, 
$\mathbf{i}$, $\mathbf{f}$, $\mathbf{o}$, $\Tilde{\mathbf{c}}$ and $\mathbf{c} \in \mathbb{R}^q$ 
denote, respectively, the input gate, forget gate, output gate, cell input activation and cell state vector;
the $\mathbf{W}$s and $\mathbf{U}$s are the weight matrices, and  the $\mathbf{b}$s are the bias terms. 
The weight matrix subscript denotes the input-output connection, e.g., $\mathbf{W}_{f}$ is the hidden-forget
gate matrix, while $\mathbf{U}_{f}$ is the input-forget matrix.

\section{Encoder-Decoder architecture}\label{sec:encdec}

Following the recent success of deep learning %
in sequence-to-sequence applications, 
we propose an encoder-decoder architecture to address the input-output mapping function~\eqref{eq:pred} %
to predict a future %
trajectory given a %
sequence of observed states and its related journey descriptor. 
A neural network architecture based on the encoder–decoder framework consists of three key components: an encoder, an aggregation function and a decoder.
The encoder reads a sequence of %
kinematic states $\mathbf{X}_\ell$ one state at a time and encodes this information %
into a sequence of hidden states. 
Then, an aggregation function  takes the hidden states computed by the encoder and produces %
a continuous-space representation $\mathbf{z}$ of the input sequence. 
Finally, the decoder generates an output sequence $\mathbf{Y}_h$ of future states step-by-step conditioned on the %
context representation $\mathbf{z}$ and the %
motion pattern descriptor $\bm{\psi}$. %
The design of the encoder and decoder may vary based on the considered task in terms of the types of neural networks employed for modelling data (e.g., CNNs, RNNs), the types of recurrent units for RNNs, the depth of the networks, etc.

In particular, we address the sequence-to-sequence prediction problem by defining an encoder-decoder architecture to learn the mapping $F_{\ell,h}$ in \eqref{eq:pred}.%

The initial encoding function $E$ in \eqref{eq:encoder}, i.e.
\begin{IEEEeqnarray}{rCl}
    \mathbf{H}_\ell &=& E( \mathbf{X}_{ \ell}; \bm{\theta}_E), \label{eq:encoder} \label{eq:ecoder}
\end{IEEEeqnarray}
is a neural network 
parameterized by $\bm{\theta}_E$,
which maps the 
input sequence 
$ \mathbf{X}_{\ell}$ 
into a sequence of internal representations $\mathbf{H}_{\ell} 
= \{\mathbf{h}_t\}_{t=1}^{\ell}$ 
such that $\mathbf{h}_{t} \in \mathbb{R}^{2q}$ is the combined hidden state of a bidirectional RNN (forward and backward recurrent networks with hidden state of size $q$). 

Then, the  autoregressive decoding function $D$ parameterized by~$\bm{\theta}_D$ in~\eqref{eq:decoder} %
predicts the future vessel state $\hat{\mathbf{y}}_{j}$ (given the previous state $\hat{\mathbf{y}}_{j-1}$) for each time step $j$, such that:
\begin{IEEEeqnarray}{rCl}
    \hat{\mathbf{y}}_j &=& D(\hat{\mathbf{y}}_{j-1},\mathbf{u}_{j},\mathbf{z}_j,\bm{\psi}; \bm{\theta}_D), \label{eq:decoder}
\end{IEEEeqnarray}
where $\mathbf{u}_{j}$ is an RNN hidden state, 
$\bm{\psi}$ is the journey descriptor,
and $\mathbf{z}_j$ is the context vector produced by a suitable aggregation function~\eqref{eq:aggregate} as follows:%
\begin{IEEEeqnarray}{rCl}
   \mathbf{z}_j &=& A( \mathbf{H}_\ell,\mathbf{u}_{j-1}; \bm{\theta}_A) \label{eq:aggregate}
\end{IEEEeqnarray}
where $A$ is able to compress the sequence of encoded hidden states $\mathbf{H}_\ell$ and
the decoder state $\mathbf{u}_{j-1} \in \mathbb{R}^{q}$ (possibly empty in the case of static aggregation functions)
into
the low-level context representation $\mathbf{z}_j$. %
The aggregation function 
generates $h$ context representations in accordance with the decoding phase. 

Finally, by 
applying sequentially 
\rev{the encoder~\eqref{eq:encoder}},
the decoder~\eqref{eq:decoder}, and the aggregation~\eqref{eq:aggregate} functions, 
the architecture generates the following output
sequence
of length $h$:
\begin{equation}
    \hat{\mathbf{Y}}_{ h}  = F_{\ell,h}(\mathbf{X}_{ \ell},\bm{\psi}),
\end{equation}
where the prediction output $\hat{\mathbf{Y}}_{h}$ represents the future %
trajectory conditioned on the sequence  $\mathbf{X}_{ \ell}$ of observed %
kinematic states and the journey descriptor $\bm{\psi}$. 

In the proposed method,
the encoder and decoder are both RNNs, which represent an end-to-end trainable architecture to address the sequence-to-sequence learning task. 
In the remainder of this section we provide more details on the encoder and decoder recurrent layers, and on how a suitable aggregation function can be used to connect together the two networks to perform the prediction task. %

\subsection{Encoder network}

The encoder network in \eqref{eq:encoder} is designed as a bidirectional RNN \cite{Schuster97}
to be trained simultaneously in positive and negative time directions.
The bidirectional architecture combines two recurrent networks, 
one processing the data in the positive temporal order, and the other in the opposite direction,
with two
separate hidden layers that are subsequently 
fed onward to the same output layer.
In particular, we use a Bidirectional LSTM (BiLSTM)~\cite{Graves2005} in~\eqref{eq:encoder} to map the input sequence $\mathbf{X}_{\ell}$ 
into two output sequences, the forward hidden sequence $\overrightarrow{\mathbf{H}}_{\ell} \triangleq 
\{ \overrightarrow{\mathbf{h}}_{t}\}_{t=1}^{\ell}$, $\overrightarrow{\mathbf{h}}_{t}\in \mathbb{R}^d$, and the backward hidden sequence $\overleftarrow{\mathbf{H}}_{\ell} \triangleq 
\{ \overleftarrow{\mathbf{h}}_{t}\}_{t=\ell}^{1}$, $ \overleftarrow{\mathbf{h}}_{t}\in \mathbb{R}^d$, 
by iterating the following operations:
\begin{IEEEeqnarray}{rCl}
\overrightarrow{\mathbf{h}}_{t} &=& \operatorname{LSTM}(\mathbf{x}_{t},\overrightarrow{\mathbf{h}}_{t-1} ;\overrightarrow{\bm{\theta}_{E}})
\label{lstm1}
\\
\overleftarrow{\mathbf{h}}_{t} &=& \operatorname{LSTM}(\mathbf{x}_{t}, \overleftarrow{\mathbf{h}}_{t+1}; \overleftarrow{\bm{\theta}_{E}}),
\label{lstm2}
\end{IEEEeqnarray}
where each $\operatorname{LSTM}$ function is a recurrent network of type \eqref{eq:LSTM}, which adapts parameters ($\overrightarrow{\bm{\theta}_E}, \overleftarrow{\bm{\theta}_E}$) to learn long-term patterns in both temporal directions.
\rev{As commonly done in BiLSTMs~\cite{Graves2005},}
the output layer of the encoder is 
finally
obtained by 
concatenating the forward and backward hidden states,
\rev{encoded by the two unidirectional LSTMs \eqref{lstm1}-\eqref{lstm2}},
into a compact \rev{bidirectional} representation 
\begin{eqnarray}\label{eq:bilstmconcat}
    \mathbf{h}_{t} = [\overrightarrow{\mathbf{h}}_{t}^\top ,  \overleftarrow{\mathbf{h}}_{t}^\top]^\top,
\end{eqnarray}
such that $\mathbf{H}_\ell$ in \eqref{eq:encoder} 
is a sequence of output vectors  $\{\mathbf{h}_1, \dots , \mathbf{h}_\ell\}$,  
where
each element $\mathbf{h}_t \in \mathbb{R}^{2q}$ encodes bidirectional spatio-temporal information extracted from the 
input vessel states preceding and following the
$t$-th component of the sequence.

\begin{figure}[!t]
    \centering
    \includegraphics[width=0.98\columnwidth]{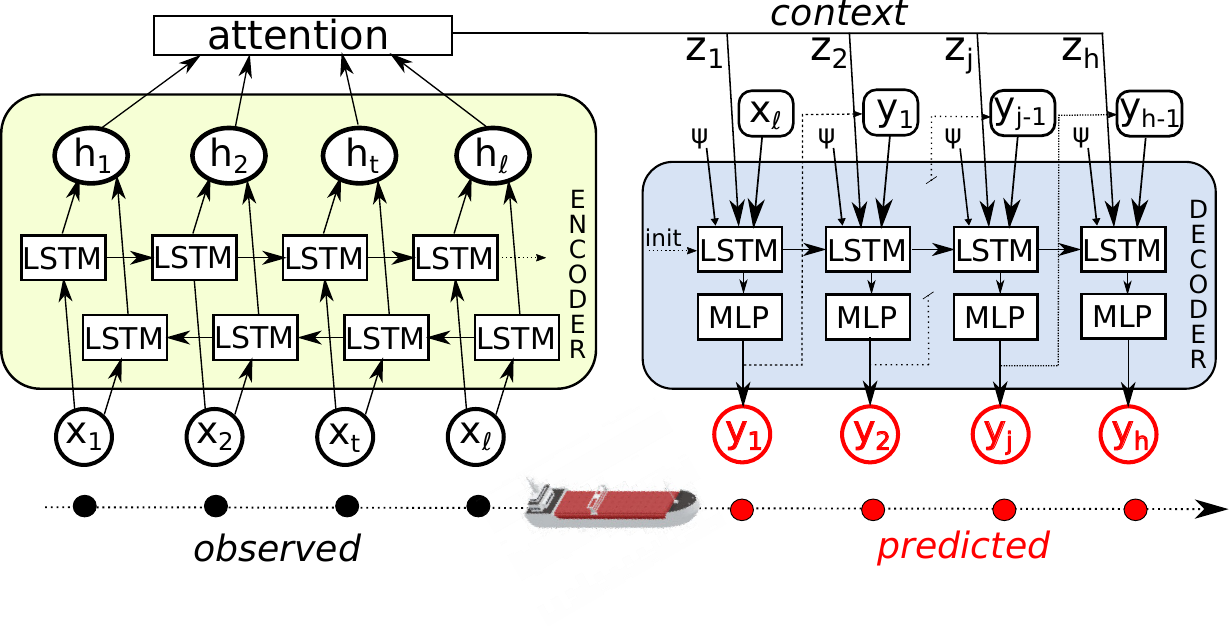}
    \caption{The proposed encoder-decoder recurrent architecture to address the task of vessel trajectory prediction. 
    The encoder is a bidirectional LSTM RNN that maps the input sequence into a sequential context representation. 
    Then, an attention-based aggregation function combines the sequence of encoder hidden states into the context representations. 
    Finally, the motion pattern descriptor and each context representation  is fed to the decoder LSTM network to generate the output sequence. 
    Encoder and decoder networks may have a multi-layered structure to improve learning of internal representations.}
    \label{fig:seq2seq}
\end{figure}
\subsection{Aggregation function}
The aggregation function $A$ in~\eqref{eq:aggregate} allows compressing all the information encoded in $\mathbf{H}_\ell$ into the contextual representation $\mathbf{z}$, %
which will be used by the decoder to produce the output sequence of length $h$. 
We can consider the aggregation as a 
pooling operation, whose
objective is to 
summarize (down-sample) the entire encoded sequence $\mathbf{H}_\ell$ in a compressed context representation $\mathbf{z}$
to preserve important information while discarding irrelevant features~\cite{Boureau2010}.
We investigate %
three possible choices of aggregation functions:

\subsubsection{Max pooling over time (MAX)}
the sequence $\mathbf{H}_\ell$ is reduced by the max-pooling operation to a single vector 
$\mathbf{z} = \operatorname{col}(z_r) \in \mathbb{R}^{2q}, r = 1, \dots,  2q$
such that
\begin{equation}
z_r = \mymax_t (\mathbf{H}_\ell)_{r,t},
\quad t \in \{ 1, \dots,  \ell\} .
\end{equation}
This type of aggregation function 
encourages capture of %
the most important feature over time~\cite{Martina2020}, 
i.e., selecting the highest unit value for each 
row $r$
over the whole temporal sequence $\mathbf{H}_\ell$. %
The resulting context vector $\mathbf{z}$ will be then repeatedly used by the decoder network to sequentially generate  the output prediction.

\subsubsection{Average pooling over time (AVG)}

the sequence $\mathbf{H}_\ell$ is reduced by the average-pooling operation to a single context vector 
$\mathbf{z} = \operatorname{col}(z_r) \in \mathbb{R}^{2q}, r = 1, \dots,  2q$
by computing 
the mean value for each hidden unit. %
In this case, each context feature  $z_r$ is computed as:
\begin{equation}
z_r = \frac{1}{\ell}\sum_{t=1}^\ell (\mathbf{H}_\ell)_{r,t}, \quad t \in \{ 1, \dots,  \ell\} .
\end{equation}
This aggregation computes an average representation of the input sequence considering the whole $\mathbf{H}_\ell$ representation computed by the encoder layer. 
The computed context representation is then repeatedly used by the decoder to produce the output prediction.

\subsubsection{Attention mechanism (ATTN)}

Attention-based neural networks have recently proven
their effectiveness in a wide range of tasks including question answering, machine translation,
and image captioning~\cite{bahdanau2015,Xu2015,Hermann2015}. 
The attention mechanism is introduced 
as an intermediate layer 
between the encoder and the decoder networks to address two issues 
that usually characterize encoder-decoder architectures using simple aggregation functions:
i) the limited capacity of a fixed-dimensional context vector regardless of the 
amount of information in the input~\cite{Cho2014} and 
ii) the lack of interpretability~\cite{bahdanau2015}.
The attention mechanism can be adopted 
to model the relation between the encoder and the decoder hidden states
and to learn the relation between the observed and the predicted kinematics states
while preserving the spatio-temporal structure of the input.
This is achieved by allowing the learnt context representation of the input  to be a sequence of fixed-size vectors, or context sequence
$\mathbf{z}= \{ \mathbf{z}_1, \dots, \mathbf{z}_{h} \}$.

Consider the sequence of encoder's hidden states  $\mathbf{H}_\ell$  
and the  hidden state $\mathbf{u}_{j-1}$ of the decoder network, 
then
each context vector $\mathbf{z}_j$ can be computed as a weighted sum of the hidden states, i.e.,
\begin{equation}
   \mathbf{z}_j =  \sum_{t=1}^\ell \alpha_{jt}\mathbf{h}_t,
\end{equation}
where $\alpha_{jt}$ represents the attention weight calculated, 
from score $e_{jt}$,
via the softmax operator as
\begin{equation}
  \alpha_{jt} = \frac{\operatorname{exp}(e_{jt})}{\sum_{t=1}^\ell \operatorname{exp}(e_{jt})},
  \label{eq:attention}
\end{equation}
with 
\begin{IEEEeqnarray}{rCl}
e_{jt} &=& \vect{v}_a^\top \operatorname{tanh}(\mathbf{W}_{h}\mathbf{h}_t + \mathbf{W}_{u}\mathbf{u}_{j-1}) .
\label{eq:combine}   
\end{IEEEeqnarray}
Note that the 
weight
matrices $\mathbf{W}_{h} \in \mathbb{R}^{2q\times q}$, 
$\mathbf{W}_{u} \in \mathbb{R}^{q\times q}$, and $\vect{v}_{a} \in \mathbb{R}^{q}$ in the trainable neural network \eqref{eq:combine} are used to combine  all encoder's hidden states $\mathbf{h}_t$ for each decoder state $\mathbf{u}_{j-1}$
to produce %
$e_{jt}$, which scores the quality of spatio-temporal relation %
between the inputs around position $t$ and the output at position $j$. %
The main task of the attention mechanism is to score 
each context vector $\mathbf{z}_j$ 
with respect to the decoder's hidden state $\mathbf{u}_{j-1}$ 
while it is generating the output trajectory 
to best approximate the predictive distribution $\eqref{eq:predictive_distr}$.
This scoring operation corresponds to assigning a probability to each context of being attended by the decoder.

\subsection{Decoder network}
The decoder $D$ in \eqref{eq:decoder} aims at generating the future trajectory given the context representation $\mathbf{z}$ of the input sequence. 
This generation phase can possibly be conditioned by the pattern descriptor $\bm{\psi}$, which, in the case of maritime traffic describes, for instance, a ship's intended journey (departure, arrival). 
The pattern descriptor augments the context information, %
helping the decoder to produce better predictions.

Formally, the autoregressive decoder is trained to predict the next vessel state $\hat{\mathbf{y}}_{j}$ given 
the context vector $\mathbf{z}_j$, 
the possibly available journey descriptor $\bm{\psi}$,  
and all the previously predicted states $\{\hat{\mathbf{y}}_{1},\dots, \hat{\mathbf{y}}_{j-1}\}$. 

\rev{Mathematically, 
the decoder computes 
the conditional probability of the predicted sequence given the observed sequence 
by factorizing the joint probability into ordered conditionals. These assume conditional independence between the past and the future sequence given the context representation and the previously predicted states, i.e.:}
\begin{equation}\label{eq:decoder2}
p(\mathbf{y}_{1}, \dots, \mathbf{y}_{h} | \mathbf{x}_{1}, \dots, \mathbf{x}_{\ell}) = \prod_{j=1}^{h} 
p(\mathbf{y}_{j}|\{\mathbf{y}_{1}, \dots, \mathbf{y}_{j-1}\},\mathbf{z}_j, \bm{\psi}).
\end{equation}
\rev{Each conditional probability in \eqref{eq:decoder2} can be modeled by an RNN $D$  and expressed as in \eqref{eq:decoder} with the following form:}
\begin{equation}
    p(\mathbf{y}_{j}|\{\mathbf{y}_{1}, \dots, \mathbf{y}_{j-1}\},\mathbf{z}_j, \bm{\psi}) = D(\hat{\mathbf{y}}_{j-1}, \mathbf{u}_{j}, \mathbf{z}_j, \bm{\psi};\bm{\theta}_{D}),
\end{equation}
where the predicted vessel state $\hat{\mathbf{y}}_{j}$ is computed by $D$ 
given 
the context vector $\mathbf{z}_j$ that summarizes %
the input sequence at the time step $j$,
the hidden state $\mathbf{u}_{j}$ of the decoder RNN, 
as well as its previous predictions $\hat{\mathbf{y}}_{j-1}$, and the journey descriptor $\bm{\psi}$. 
The advantage of a decoder network with such a recursive architecture 
is that it can, by design, model sequences of arbitrary length. In our case, we consider an output sequence of length $h$, and we implement the decoder as a unidirectional LSTM network, 
which generates the sequence of future predictions by iterating for $j = 1, \dots, h$ 
a single forward layer as follows %
\begin{IEEEeqnarray}{rCl}
\bm{\mu}_t &=& [\hat{\mathbf{y}}_{j-1}, \mathbf{z}_j, \bm{\psi}]\\
\mathbf{u}_j &=& \operatorname{LSTM}(\bm{\mu}_t, \mathbf{u}_{j-1}; \bm{\theta}_D)    \\
\hat{\mathbf{y}}_{j} &=& \mathbf{W}_{y} \mathbf{u}_{j} + \mathbf{b}_{y},
\end{IEEEeqnarray}
where $\hat{\mathbf{y}}_0 = \mathbf{x}_\ell$, 
$\mathbf{z}_j$ is the context vector computed by the aggregation function, 
$\mathbf{W}_{y}$ and $\mathbf{b}_{y}$ are trainable parameters of a neural network that maps the LSTM output $\mathbf{u}_j \in \mathbb{R}^q$ into the next predicted state $\hat{\mathbf{y}}_{t}$. 

Note that the first operation is to concatenate the previous predicted state $\hat{\mathbf{y}}_{j-1}$, 
the context information $\mathbf{z}_j$, and the one-hot descriptor $\bm{\psi}$ into the input vector $\bm{\mu}_t$. The dimension of $\bm{\mu}_j$ is dependent on the nature of the aggregation function, so that when a static aggregation function (MAX or AVG) is used, then $\mathbf{z}_j \in \mathbb{R}^{2q}$, otherwise $\mathbf{z}_j \in \mathbb{R}^{q}$.

In~\cite{bahdanau2015} the authors propose to initialize the decoder hidden states with the last backward encoder state. %
We take a slightly different path and initialize the decoder state with 
\begin{equation}
\mathbf{u}_0 = \operatorname{tanh}(\mathbf{W}_{\kappa} \overrightarrow{\mathbf{h}}_{\ell} + \mathbf{b}_{\kappa}),
\label{eq:decoder_init}
\end{equation}
where $\mathbf{W}_{\kappa} \in \mathbb{R}^{q \times q}$ and $\mathbf{b}_{\kappa}$ are trainable parameters.%

This end-to-end solution is trained by the stochastic gradient descent algorithm in order to learn an optimal function approximation \eqref{eq:pred}.
Figure~\ref{fig:seq2seq} illustrates the proposed sequence-to-sequence architecture 
for trajectory prediction
with the attention-based aggregation function as intermediate layer between encoder and decoder networks.

\section{Experimental setup}
\label{sec:experiments}

In this section we 
describe how 
to apply the proposed method 
to %
vessel trajectory prediction using real-world AIS data. 
To this aim, we assemble %
a dataset that will serve as benchmark 
to evaluate 
different deep-learning solutions
in terms of prediction performance.

\subsection{Dataset preparation}

\begin{figure*}
    \centering%
    \includegraphics[width=.98\textwidth]{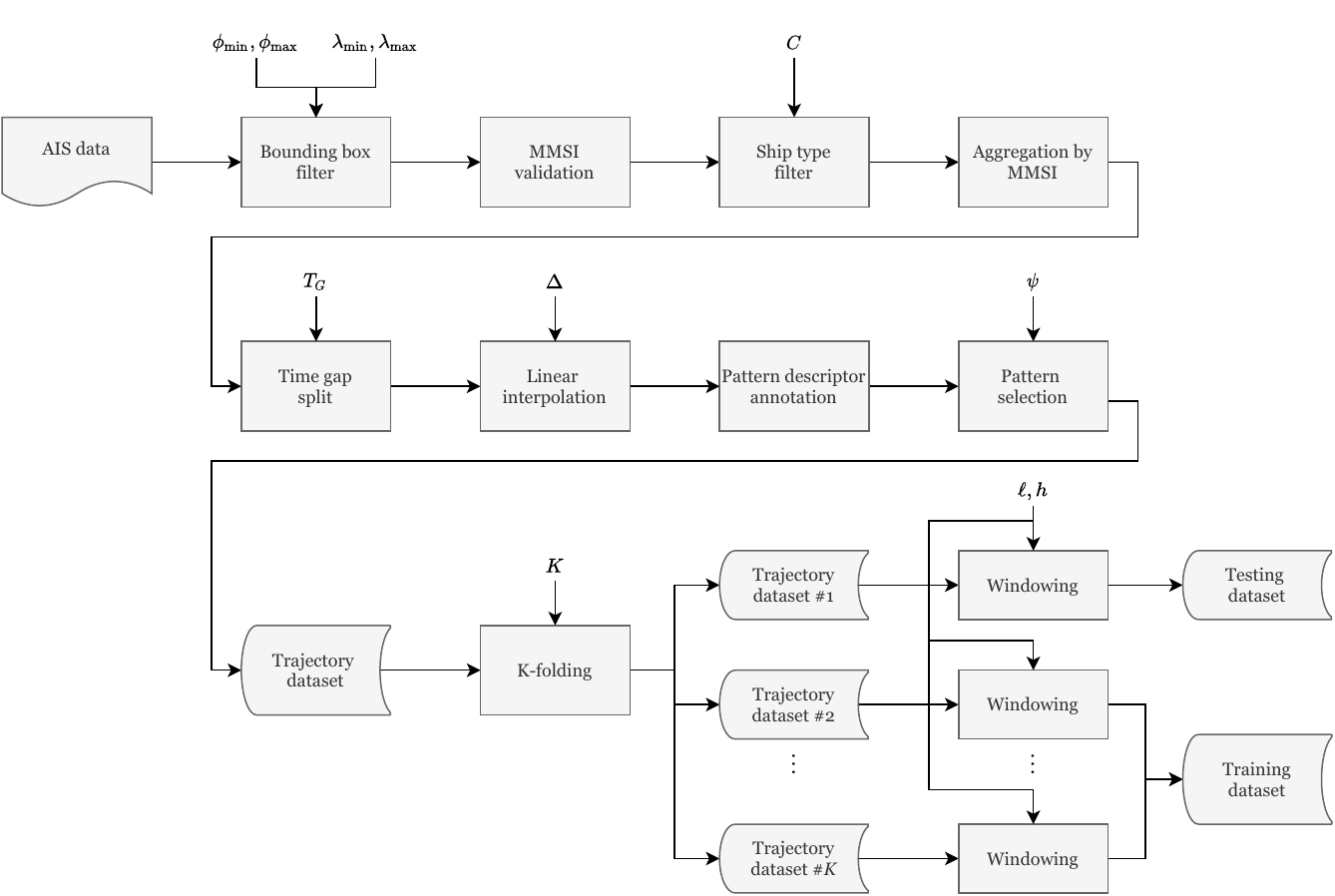}%
    \caption{\rev{Flowchart of the dataset preparation method used in the experiments.}}
    \label{fig:diagram}
\end{figure*}

\rev{A flowchart of the dataset preparation routine is illustrated in Fig.~\ref{fig:diagram}. The input is represented by AIS data in tabular format~\cite{AIS_ITU}; each row in the input dataset corresponds to a positional AIS message broadcast by a ship and must contain: a timestamp, the ship's identification (MMSI number), the ship's position (latitude, longitude coordinate pair), and the ship type (cargo, tanker, passenger, etc.) As a first step, we apply a bounding box filter, which retains only the AIS messages broadcast from within a rectangular bounding box defined by its latitude and longitude extents. Then, a MMSI validation step discards the AIS messages with invalid MMSI numbers (i.e., numbers with less than nine digits and numbers whose first three digits do not correspond to a valid MID code). Subsequently, we retain only messages broadcast by a specific ship type $C$. After this step, AIS messages can be aggregated by MMSI; the aggregation transforms the dataset of positions into a collection of trajectories, i.e., time-ordered lists of messages broadcast by a single vessel, identified with its MMSI number. Each trajectory is then split into one or more sub-trajectories, so that the time intervals between subsequent AIS messages in every sub-trajectory are all below a given threshold $T_G$. It follows a linear interpolation step, which resamples the data at a fixed interval of $\Delta$ seconds. After the interpolation, each trajectory is (possibly) annotated with a pattern descriptor and only the trajectories belonging to a specific set of patterns $\Psi$ are retained; the output of this step is a trajectory dataset that will be persisted and will undergo a $K$-fold splitting procedure, which splits the dataset into $K$ groups with approximately the same number of trajectories. Finally, a windowing procedure generates, from the trajectory data, a list of input/target sequences of length $\ell$ and $h$, respectively, that will be used to train and test the models.}

\subsection{\rev{Scenario description}}
The data that we used in our experimental setup comes from the Danish Maritime Authority (DMA), which makes 
\ac{AIS} data from Danish waters freely available~\cite{DMA}.
As evident from the density map in Fig.~\ref{fig:emodnet}, the area of the analysis is %
characterized by complex maritime traffic patterns, which may result difficult to analyze.
For this reason, we focus only on \ac{AIS} positions of \textit{tanker} vessels, both for computational reasons and to create a dataset as homogeneous as possible; the time period used for our analysis spans from January to February 2020.

More specifically, we focus our attention only on two motion patterns %
defined by a \textit{departure} area, identified by the polygon $O$ in Fig.~\ref{fig:emodnet}, and two \textit{destination} areas, identified by the polygons $A$ and $B$ in Fig.~\ref{fig:emodnet}. Given the polygons $O$, $A$ and $B$, we build the dataset of \ac{AIS} trajectories limited to those that transit %
in $O$ and, subsequently, $A$ or $B$; in other words, we build the set $\mathcal{C}$ described in Section~\ref{sec:sec2_1}, where the categorical feature $\Psi$ represents, in this case, the destination ($A$ or $B$).

\rev{This selection stage leaves us with a dataset of trajectories with variable duration---as much as thirteen hours;}
the time interval between consecutive reported positions is also variable and has a mean value of about $10$ seconds. %
At this point, the dataset is ready to be resampled, and the trajectories are all interpolated using a
fixed sampling time. In our experimental setup, we selected a sampling time of 
$\Delta = 15$ minutes, as a good trade-off between training time and time resolution, considering the limited maneuverability of tanker vessels. %
The final dataset contains %
$394$ trajectories, out of which $232$ belong to the motion pattern $(O,A)$, 
and $162$ to $(O,B)$.  
Finally, before starting the experiments, we perform the data windowing and segment trajectories into fixed-length input/output %
sequences of length $\ell=h=12$ (equivalent to three hours, with a sampling period of $\Delta=15$ minutes) that we will use to train and evaluate the models (see Section~\ref{se:segmentation}). \rev{As already mentioned, the windowing procedure produces a dataset of input/target sequences, which are of fixed lengths ($\ell$ and $h$, respectively); the exact number of sample sequences is reported in~Section\ref{sec:qualitative}.}

\subsection{Models}
We compare the performance of the developed encoder-decoder architecture %
for different aggregation mechanisms 
against baseline methods that can be applied to trajectory prediction problems. We test different
deep-learning models, %
whose hyper-parameters have been selected after an extensive hyper-parameter optimization procedure aimed at finding
the most %
suitable configurations. A list of models evaluated in our analysis follows.
\begin{itemize}
    \item \textit{Linear regression (Linear)}: 
    The multi-output linear regression model~\cite{Gareth2014} predicts %
    the future trajectory given the full motion history. Each predicted value in the output trajectory is computed by an independent linear regressor.%
    
    \item \textit{Multi Layer Perceptron (MLP)}: A feed-forward neural network~\cite{Goodfellow2016} that maps %
    a given input sequence into a future output sequence~\eqref{eq:pred}. In our setup, the model receives the input sequence of $\ell=12$ pairs of latitude, longitude values as a flattened vector. The network is composed of two  hidden layers with $512$ neurons, both followed by the ReLu activation~\cite{Glorot11}. Then, the final layer computes a linear transformation from the hidden features to $24$ output units, which corresponds to $12$ prediction states (longitude and latitude), a duration of 3 hours, %
    given a sampling period of $\Delta=15$ minutes. %
    We train the MLP~\eqref{eq:lossmin} %
    with the mean absolute error loss function. The choice of this specific loss function %
    is a result of preliminary experiments %
    varying the network's width and depth, which showed that the mean absolute error allows achieving 
    the best performance. %
    Adding more layers or augmenting  the network width lead to 
    weaker performance (stronger over-fitting).

    \item \textit{Encoder-Decoder (EncDec)}: This is the proposed architecture described in Section~\ref{sec:encdec}, which is composed of a BiLSTM encoder layer with $64$ hidden units, and a LSTM decoder layer with $64$ hidden units. 
    We compare three different aggregation functions: Max global pooling (MAX), Average global pooling (AVG) and the Attention mechanism (ATTN). 
    We train the model~\eqref{eq:lossmin} by using the mean absolute error loss function, which we selected again as the one that achieved the best performance in preliminary experiments. %
\end{itemize}

Moreover, inspired by~\cite{Ristic08}, we compare two fundamentally different approaches to account for the vessel behavior
based on labeled and unlabeled trajectories. %
In other words, in the \textit{labeled} case the models are trained to exploit,
when available,
the high-level pattern information
 $\bm{\psi}$ that represents the vessel's intended destination. %
 Therefore, we evaluate an \textit{unlabeled} and a \textit{labeled} version for each of the three models in the previous list. %
\begin{itemize}
    \item \textit{Unlabeled (U)}: We train the predictive model and perform prediction using unlabeled trajectories, i.e., 
    using only
    low-level context representation encoded from 
    a sequence of 
    past observations,
    without any
    high-level
    information about the motion intention. 
    The employed model is that described in Section~\ref{sec:encdec}, where the pattern descriptor $\bm{\psi}$ is removed from the decoding phase.
    This amounts to considering a reduced %
    version of the proposed architecture 
    to compute~\eqref{eq:pred}. %
    \item \textit{Labeled (L)}: We train the model and perform prediction using labeled trajectories,  %
    i.e.,
    using
    low-level context representation encoded from 
    a sequence of 
    past observations, 
    as well as
    additional inputs $\bm{\psi}$
    about high-level intention behavior of the vessel (final destination).
    In our experiments, the descriptor defines the %
    destination area, i.e., one of the polygons $A$ or $B$ depicted in Fig.~\ref{fig:density_maps}. %
    This model %
    exploits the descriptor in the generating phase to predict the future trajectory. 
    For the linear and MLP models, the descriptor is incorporated directly %
    into the input vector.
\end{itemize}

\subsection{Training settings}

The final goal of the learning process
is to find the best approximation for the predictive function, i.e., the function that maps past trajectories into future ones,  %
both composed of continuous values. 
First, we apply a linear transformation to latitude and longitude pairs %
so that the data, after the transformation, is distributed with
with zero mean and unit variance. 
Then, we train the neural networks %
with ADAM~\cite{Kingma2014}, a gradient-based algorithm to minimize the loss function and learn the internal parameters. We use the ADAM default configuration proposed in
\cite{Kingma2014} 
where $\beta_1=0.9$, $\beta_2=0.999$, and $\epsilon=10^{-8}$. %
We let the training run for $3000$ epochs max
with a learning rate of $0.0001$ %
and a mini-batch size of $200$ samples;  
we use the early-stopping criterion evaluating the error measure (mean absolute error between prediction and ground truth sequences) on the validation set and providing guidance on how many iterations we can perform before the model begins to over-fit \cite{Goodfellow16}.%

Initialization is a crucial step to achieve satisfactory performance.
We initialize the MLP weights by using the He method~\cite{Kaiming2015}. %
The LSTM-based encoder-decoder architecture needs a different initialization, 
so we follow recent approaches using orthogonal
initializations of the transition
matrices for LSTM RNNs~\cite{Henaff2016, Vorontsov2017}, 
while the other weight matrices are initialized using the Xavier schema~\cite{Glorot2010}.  Moreover, we initialize the biases of the forget gate to $1$, and all other gates to $0$~\cite{Rafal2015}.

\subsection{Evaluation  metrics}
In vehicle trajectory prediction, %
the model performance is usually assessed with %
the average displacement error (ADE)~\cite{Pellegrini2009} on all the predicted positions with respect to ground truth. 
This metric 
measures
the performance in terms of mean square error (MSE) between the actual an the predicted positions.
In our specific application, 
\rev{the positions being expressed}
as pairs of geographical coordinates,
the prediction error $d_{H}$ is defined as
the \textit{great-circle} distance between the true and the predicted positions on the Earth surface. 
We use the \textit{haversine formula} to calculate the distance between any two points $\mathbf{s}_1$ and $\mathbf{s}_2$ expressed by their geographical coordinates: %
\begin{equation}\label{eq:haversine}
    d_{H}(\mathbf{s}_1, \mathbf{s}_2) = 2R \arcsin \sqrt{\sin^2 \tilde{\phi}+ \cos\phi_1\cos\phi_2 \sin^2 \tilde{\lambda}},
\end{equation}
where 
\rev{$\mathbf{s}_i = [\phi_i, \lambda_i]^T, \, i=1,2$,}
$R$ is the radius of the Earth, 
$\phi_1$ and $\phi_2$ denote the latitude values, $\lambda_1$ and $\lambda_2$ the longitude values,
$\tilde{\phi}=\frac{\phi_2-\phi_1}{2}$, and $\tilde{\lambda}=\frac{\lambda_2 - \lambda_1}{2}$.

Considering the haversine formula~\eqref{eq:haversine} as a distance function expressed in nautical miles (nmi), %
the \textit{mean absolute error} (MAE) can be defined as the displacement haversine error over a set of $N$ test samples, i.e.,
\begin{equation}
  \operatorname{MAE}_j = \frac{1}{N} \sum_{i=1}^N d_{H}(\mathbf{y}_{j}^i, \hat{\mathbf{y}}_{j}^i)
\end{equation}
where $\mathbf{y}_{j}^i$ and $\hat{\mathbf{y}}_{j}^i$ represent the true position and, respectively, the predicted position of sample $i$ after a given prediction horizon 
$t_j=j \, \Delta$, $j=1, \dots, h$.%

\textit{Remark:} Even if the area of the experiment is limited, the difference between the great-circle and the Euclidean distance in the longitude, latitude space is not always negligible. Especially over a waypoint and for long time horizons, the predicted position might be significantly off from the ground truth; in these cases, the great-circle distance provides a more correct estimation of the prediction error.

\begin{table}
\renewcommand{\arraystretch}{1.3}
\centering%
\caption{Comparison of the MAE (expressed in nautical miles) for different prediction horizons in a 5-fold cross validation study achieved with unlabeled and labeled models. In parenthesis, we report the percent improvement of labeled models with respect to their unlabeled counterpart.}
\label{tbl:compare}
\resizebox{\columnwidth}{!}{%
\begin{tabular}{cc|ccc|ccc}

& & \multicolumn{3}{c|}{Unlabeled} & \multicolumn{3}{c}{Labeled}\\
{\footnotesize Model} & {\footnotesize Aggr.} & {\footnotesize 1h} & {\footnotesize 2h} & {\footnotesize 3h} & {\footnotesize 1h} & {\footnotesize 2h} & {\footnotesize 3h} \\\hline
Linear & -    & \tablenum{1.51} & \tablenum{3.99} & \tablenum{7.12} & \tablenum{1.48} \hfill(\SI{6 }{\percent}) & \tablenum{3.74} \hfill(\SI{6 }{\percent}) & \tablenum{6.14} \hfill(\SI{14}{\percent}) \\
MLP    & -    & \tablenum{0.89} & \tablenum{2.13} & \tablenum{4.01} & \tablenum{0.67} \hfill(\SI{25}{\percent}) & \tablenum{1.24} \hfill(\SI{42}{\percent}) & \tablenum{1.96} \hfill(\SI{51}{\percent}) \\
EncDec & MAX  & \tablenum{0.81} & \tablenum{2.01} & \tablenum{3.80} & \tablenum{0.60} \hfill(\SI{26}{\percent}) & \tablenum{1.16} \hfill(\SI{42}{\percent}) & \tablenum{1.84} \hfill(\SI{52}{\percent}) \\
EncDec & AVG  & \tablenum{0.80} & \tablenum{2.02} & \tablenum{3.84} & \tablenum{0.56} \hfill(\SI{30}{\percent}) & \tablenum{1.10} \hfill(\SI{45}{\percent}) & \tablenum{1.77} \hfill(\SI{54}{\percent}) \\
EncDec & ATTN & {\bfseries \tablenum[detect-weight=true]{0.78}} & {\bfseries \tablenum[detect-weight=true]{1.93}} & {\bfseries \tablenum[detect-weight=true]{3.66}} & {\bfseries \tablenum[detect-weight=true]{0.56}} \hfill(\SI{28}{\percent}) & {\bfseries \tablenum[detect-weight=true]{1.08}} \hfill(\SI{44}{\percent}) & {\bfseries \tablenum[detect-weight=true]{1.73}} \hfill(\SI{53}{\percent}) \\
\end{tabular}%
}
\end{table}

\subsection{Results: quantitative  analysis}

We use the $K$-fold cross validation procedure
to evaluate the proposed methods on a limited data sample~\cite{Hastie2009}. This method involves randomly dividing the set of trajectories into $K$ groups of approximately equal size. %
In our experiments, we split the dataset composed of $394$ full trajectories in $K=5$ folds. 
For each split, one fold is used for testing, 
while the remaining $K-1$ folds are used to fit and validate the model in the training phase. 
In this way, we can have complete and independent trajectories in each fold. \rev{The windowing procedure generates approximately %
\num{2400} samples of fixed-length input/target sequences in each fold.
Note that the exact number of sample sequences varies with the length of the trajectories falling into a specific fold, which is variable, because the windowing procedure applied to the generic $i$-th trajectory generates $n_i$ samples, with $n_i\gg \ell, h$. usually} Finally, we compute the average MAE over \rev{the extracted sequences from the} five folds,
obtaining the results shown in Table~\ref{tbl:compare}. 
As we can see from Table~\ref{tbl:compare},
the encoder-decoder architecture (EncDec) achieves the best performance for each considered prediction horizon 
in both the %
labeled and unlabeled %
experiments.
The ATTN aggregation mechanism (see Section~\ref{sec:encdec})
outperforms both the MAX and AVG aggregation functions; intuitively, the reason of this behavior can be found in the increased flexibility of ATTN~\eqref{eq:combine} to learn spatio-temporal relationships among \textit{all} input states and output states.

Comparing the results obtained with %
labeled and unlabeled training data, %
we can see that deep learning-based models are able to leverage the  high-level pattern descriptor better  and %
achieve better prediction, as opposed to linear regression. %
Moreover, the MAE of labeled neural models with prediction horizon of $3$ hours is shown to be around half of the error obtained using unlabeled input data. 
We also note that the high-level pattern information is especially important for long-term prediction. Indeed, while both unlabeled EncDec and MLP achieve reasonable prediction performance in the short term, for longer time horizons, not using the motion pattern descriptor makes the prediction performance degenerate quickly. %

\subsection{Results: qualitative analysis}\label{sec:qualitative}

\begin{figure}[!t]
    \centering%
    \subfloat[][]{%
        \includegraphics[trim=90 10 130 10,clip,width=.49\textwidth]{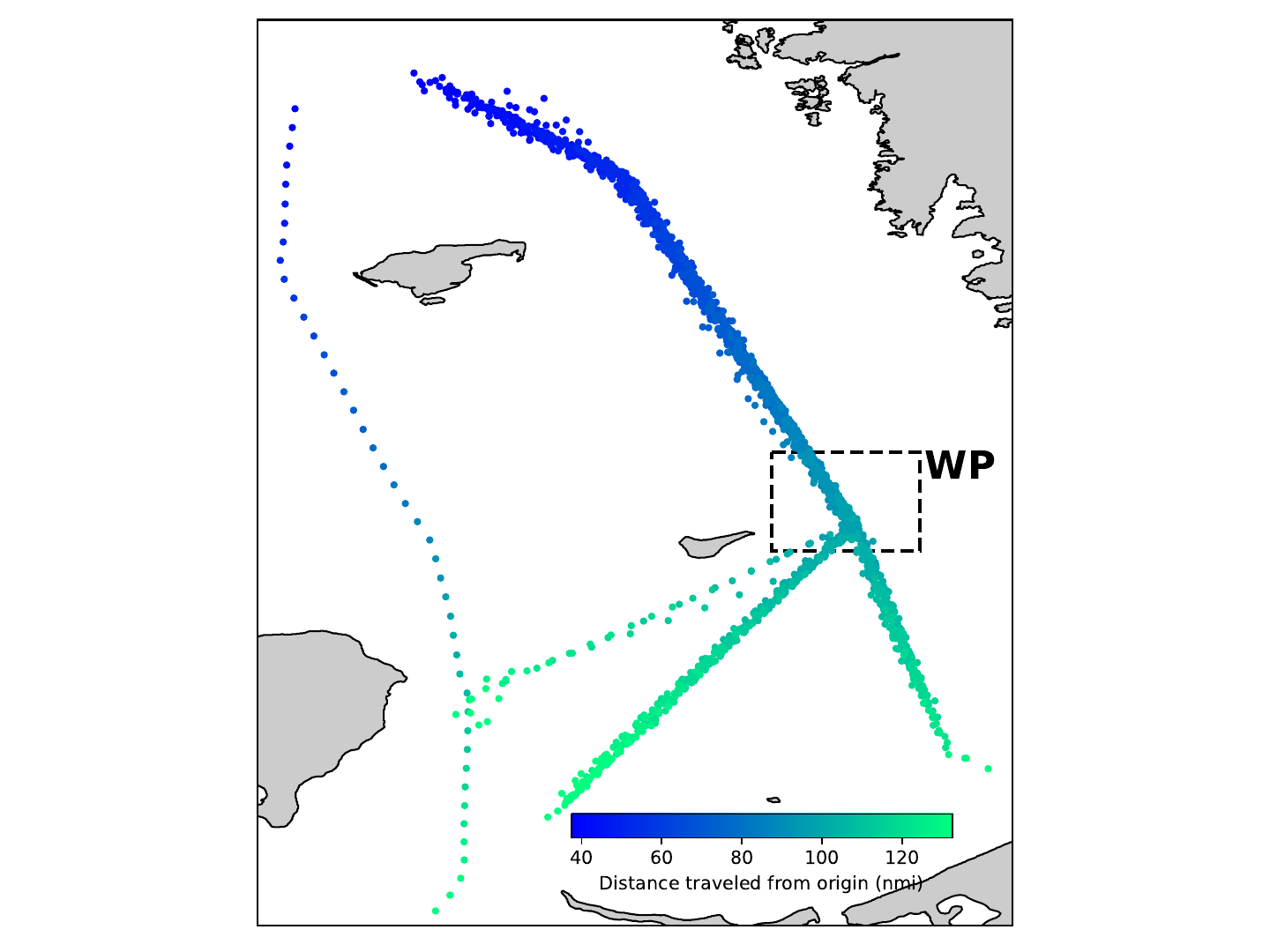}%
        \label{fig:compare_dist_b}
        }%
    \hfil
    \subfloat[][]{%
       \includegraphics[trim=5 10 0 10,clip,width=.49\textwidth]{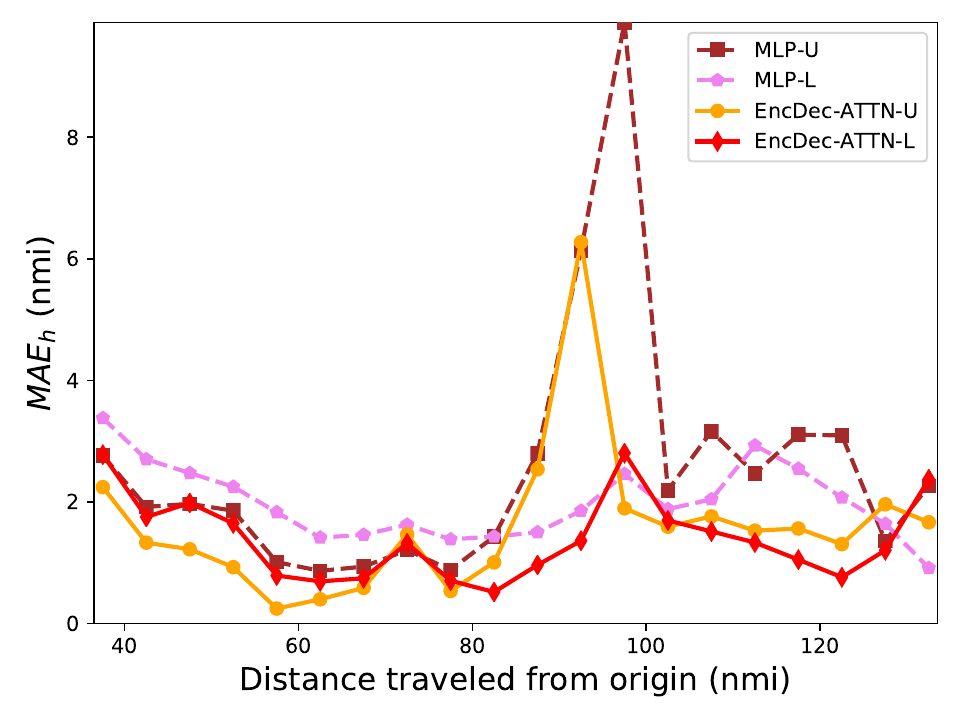}%
        \label{fig:compare_dist_a}%
        }%
    \caption{$\operatorname{MAE}_j$ for $j=h$ (i.e., with prediction horizon of $3$ hours) as a function of the vessel distance from a fixed origin point. The prediction error of the MLP model and the EncDec models are compared for a network with or without the ship's intended journey descriptor as input information. %
    \rev{Contains data from the Danish Maritime Authority that is used in accordance with the conditions for the use of Danish public data~\cite{DMA}.}
    }%
    \label{fig:compare_distance}%
\end{figure}
In this section, we provide a \textit{qualitative} analysis of the prediction results by fixing the $K$-fold split index, \rev{which translates to using a dataset with} %
\num{284} trajectories for training, \num{32} for the validation, and \num{78} for the test. \rev{The windowing procedure (see Section~\ref{se:segmentation})  produces \num{8574} input/target sequences with length $12$ samples each ($\ell=h=12$) for training and \num{1054} input/target sequences for validation. Then, we use \num{2379} input/target sequences in the testing phase to evaluate the model performance.}%
\begin{figure}[t]
    \centering
     \includegraphics[trim=10 190 10 190,clip, width=\columnwidth]{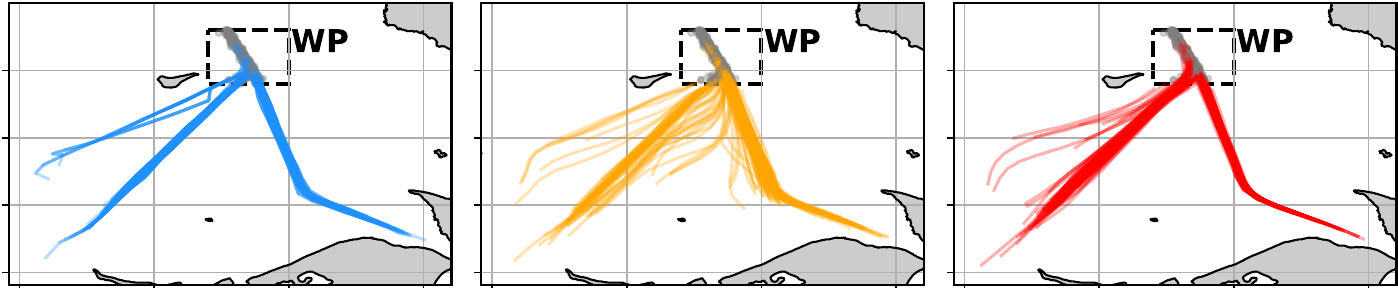}
    \caption{Comparison of computed predictions versus ground-truth vessel trajectories in a pre-determined water zone. From left to right, the plots depict: the real target trajectory, the EncDec-ATTN unlabeled predictions, and the EncDec-ATTN labeled prediction. \rev{Contains data from the Danish Maritime Authority that is used in accordance with the conditions for the use of Danish public data~\cite{DMA}.}}
    \label{fig:prediction_distribution}
\end{figure}

The test trajectories of the chosen split are illustrated in Fig.~\ref{fig:compare_dist_b}, with a color that changes with the distance traveled from a fixed origin point. Figure~\ref{fig:compare_dist_a} illustrates the $\operatorname{MAE}_{h}$ as a function of the vessel's 
traveled distance
from the fixed origin point. 
Intuitively, this allows mapping all the prediction errors on a common axis and identifying 
the regions where the prediction error is high.
In terms of prediction error, the EncDec with ATTN outperforms the MLP model for each value of traveled distance. %
On the other hand, 
\rev{a significantly lower prediction error}
can be %
achieved with the proposed EncDec using labeled training data. %
In particular, we can see that %
the $\operatorname{MAE}_{h}$ obtained with unlabeled models is very high compared to that of the EncDec labeled model in a specific geographical area $WP$ (distance between $80$ and $100$ nmi), which is also highlighted in Fig.~ \ref{fig:compare_dist_b}. 
Interestingly, a branching is present in this area, where vessels can take %
three different directions; it is therefore very challenging for the prediction system to decide ahead which path the vessel is going to take.  
It is also worth noting that 
high-level information is key to improve prediction performance corresponding to branching areas (e.g., area $WP$ in Fig. \ref{fig:compare_dist_b}), while, outside such areas, the labeled models tend to put heavy weight on (or co-adapt too much to)
the journey descriptor, and not enough on 
the low-level context representation (past observed states).  
The major difference between unlabeled and labeled predictions can be explained as follows: the former %
use the past observed positions to generate future states, while the latter %
also use an additional high-level pattern information, i.e., %
the ship's intended destination: in our experiments, the polygons $A$ or $B$, which can be replaced, for instance, by the destination port in practical applications. %
In the area $WP$ (Fig.~\ref{fig:compare_dist_b}), the main disadvantage of the unlabeled models is the unavailability of any prior information that helps them to %
take the correct decision about 
the future path to be followed by the vessel after branching. %
Instead, the labeled model has more information available, and thus generates more realistic future trajectories by exploiting the information related to the pattern descriptor. %
As clearly shown in Fig.~\ref{fig:prediction_distribution},
unimodal trajectory prediction
models which are independent of the vessel's intended journey 
may tend to average out all the possible trajectory modes,
since the average prediction %
minimizes the displacement error.
However, the average of modes
may not necessarily represent a valid future behavior of the vessel. %
This leads, in the case of unlabeled model, to the generation of 
future trajectories that may have never been seen in the training phase.
Such \textit{averaging} effect of the unlabeled predictions will result in predicted trajectories  
with higher variability (see Fig. \ref{fig:prediction_distribution}) with respect to the target future positions.
The labeled model, instead, generates future trajectories that are consistently more similar to the real (target) ones,
as we can see from Fig.~\ref{fig:prediction_distribution} showing the target and predicted trajectories from the labeled and unlabeled EncDec-ATTN models. 

\begin{figure}[!t]
    \centering
    \includegraphics[width=0.98\columnwidth]{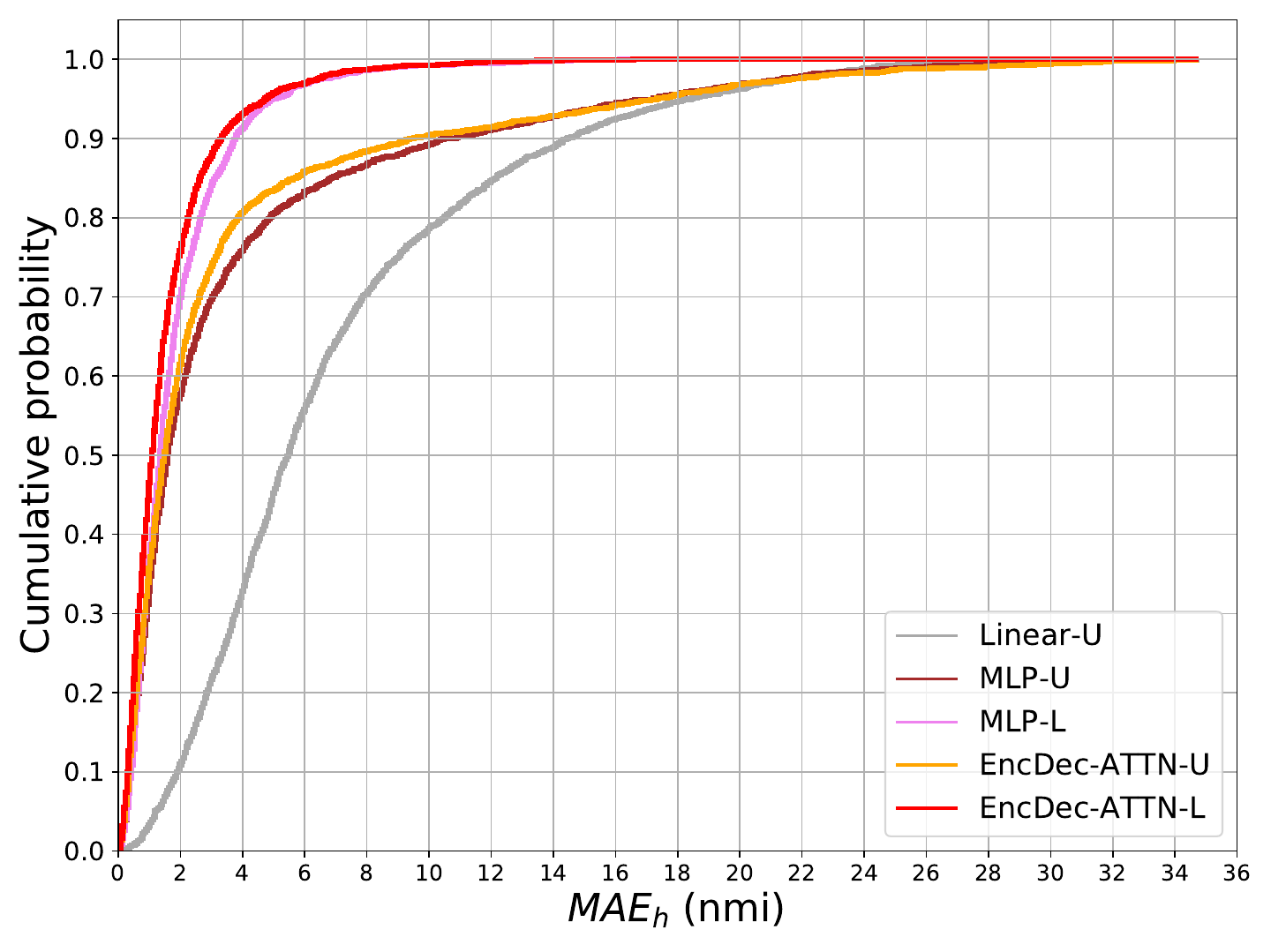}
    \caption{Empirical cumulative density function of the mean absolute error (with prediction horizon of $3$ hours) in nautical miles. We compare \rev{five} models, showing that the labeled models outperform the unlabeled models in terms of the probability that $\operatorname{MAE}_h$ will take on a value less than or equal to a specific haversine distance.}%
    \label{fig:cdf}
\end{figure}

In Fig.~\ref{fig:cdf} we compare the empirical cumulative distributive function (CDF) of the prediction error in terms of $\operatorname{MAE}_h$ for different models. This figure illustrates the complete distribution of the prediction error and complements the information already conveyed by Table~\ref{tbl:compare} and Fig.~\ref{fig:compare_distance}, showing that the labeled prediction consistently outperforms the unlabeled one. 
Note that the labeled EncDec model with attention aggregation achieves a final displacement error below \num{2.5} nmi in approximately $90\%$ of the cases. %
Considering the unlabeled models, we can see that the EncDec is slightly better than the MLP, and that the accuracy in prediction after \num{3} hours of the neural architectures (MLP and EncDec) is lower than \num{2.5} nmi in approximately $70\%$ of the cases. \rev{Considering also the labeled models, we note that EncDec is slightly better than the MLP, confirming that the intention information is a key factor to increase the prediction performances of different neural architectures.} %
Conversely, the linear regression model is not able to achieve comparable prediction performance compared to the considered deep-learning solutions.

\begin{figure*}
\centering %
\subfloat[$k-1$]{%
    \includegraphics[trim=35 10 35 10,clip,width=.3\textwidth]{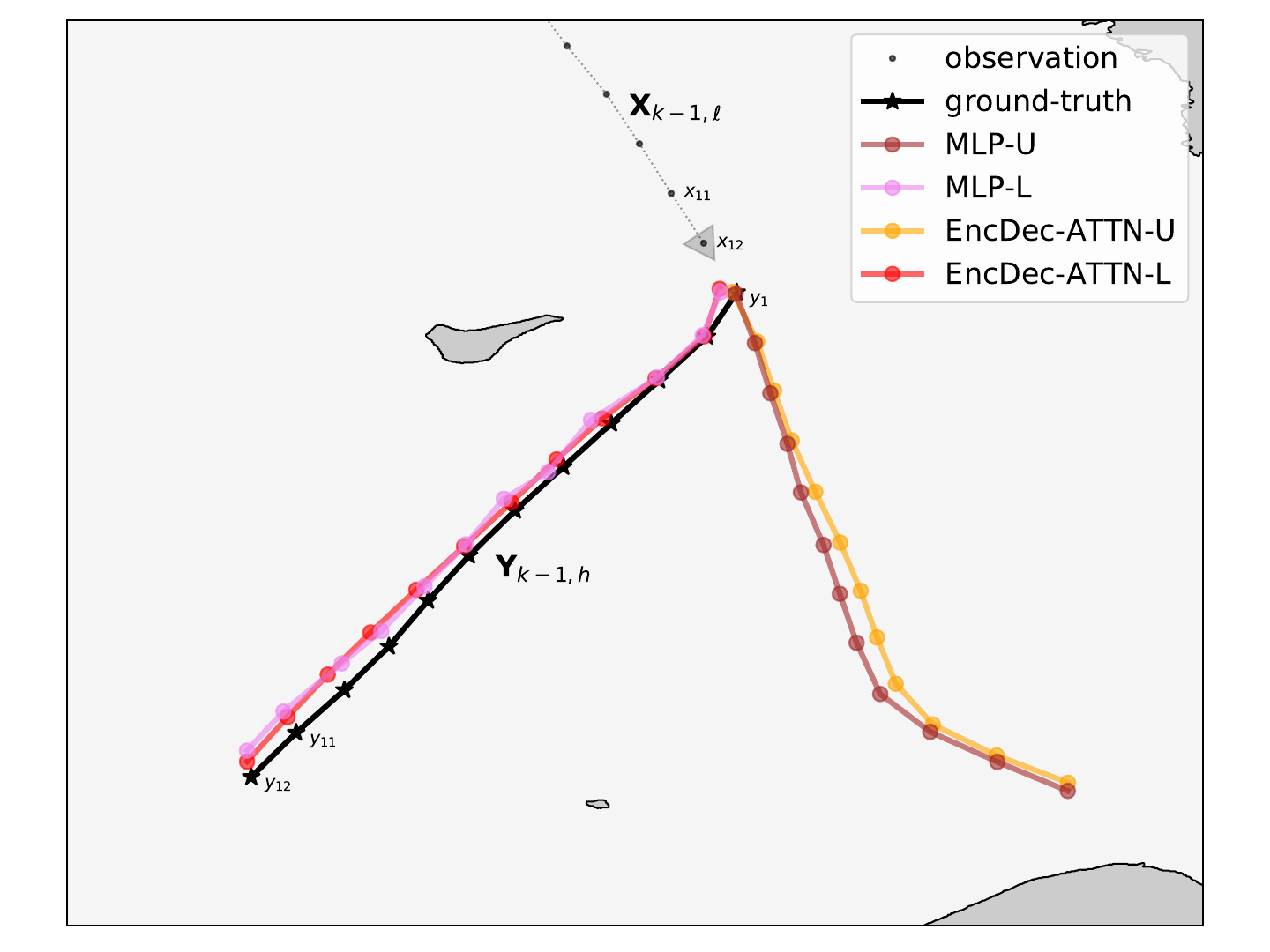}%
    \label{fig:prediction1}} \hfill
\subfloat[EncDec-ATTN-U ($k-1$)]{%
    \includegraphics[trim=50 20 50 35,clip,width=.3\textwidth]{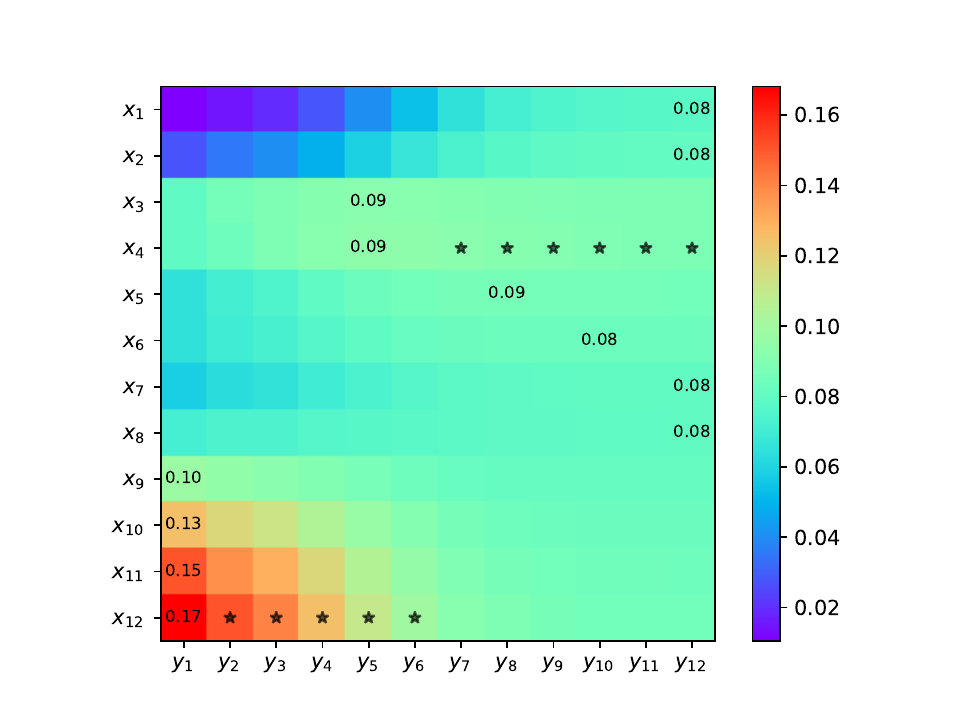}%
    \label{fig:attn_u_1}} \hfill
\subfloat[EncDec-ATTN-L ($k-1$)]{%
    \includegraphics[trim=50 20 50 35,clip,width=.3\textwidth]{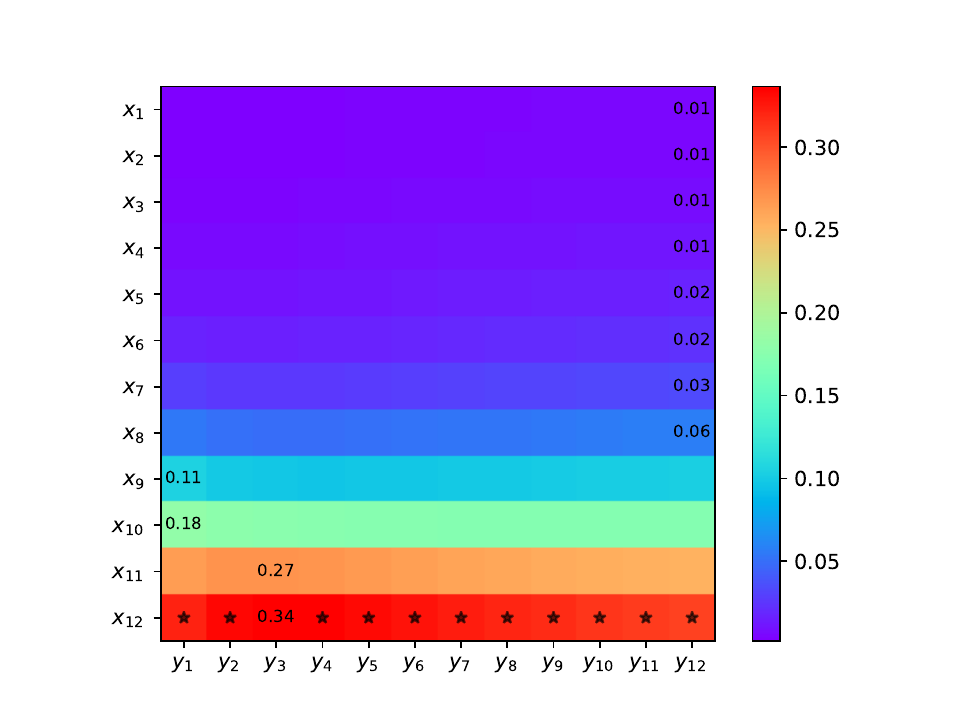}%
    \label{fig:attn_l_1}} \\
\subfloat[$k$]{%
    \includegraphics[trim=35 10 35 10,clip,width=.3\textwidth]{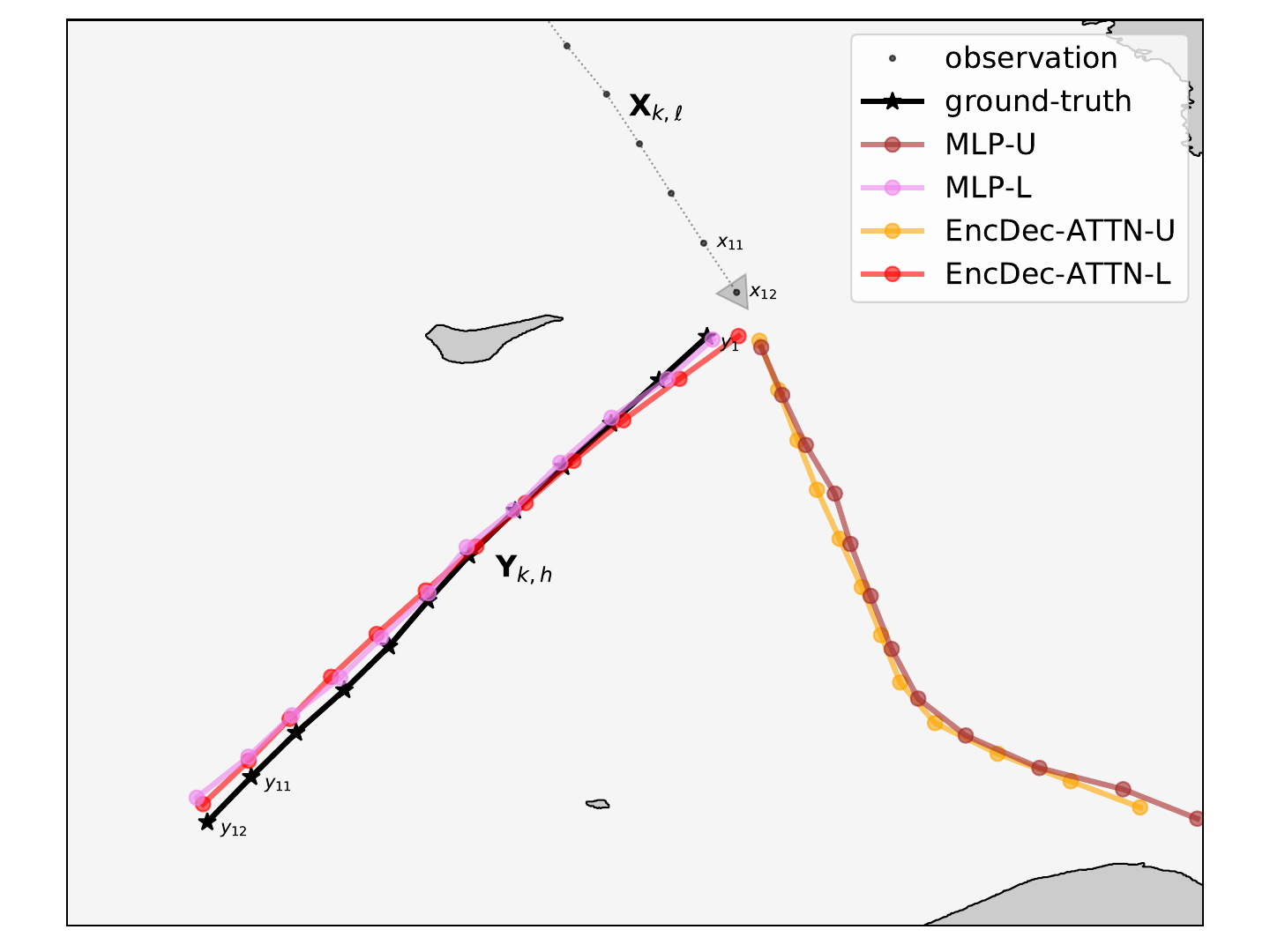}%
    \label{fig:prediction2}} \hfill
\subfloat[EncDec-ATTN-U ($k$)]{%
    \includegraphics[trim=50 20 50 35,clip,width=.3\textwidth]{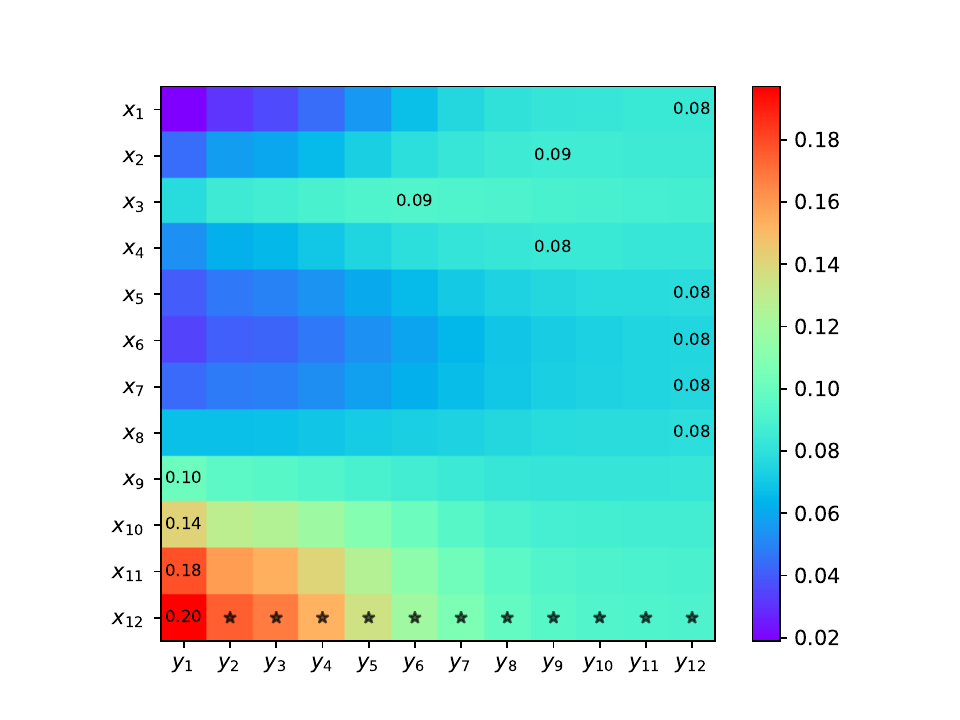}%
    \label{fig:attn_u_2}} \hfill
\subfloat[EncDec-ATTN-L ($k$)]{%
    \includegraphics[trim=50 20 50 35,clip,width=.3\textwidth]{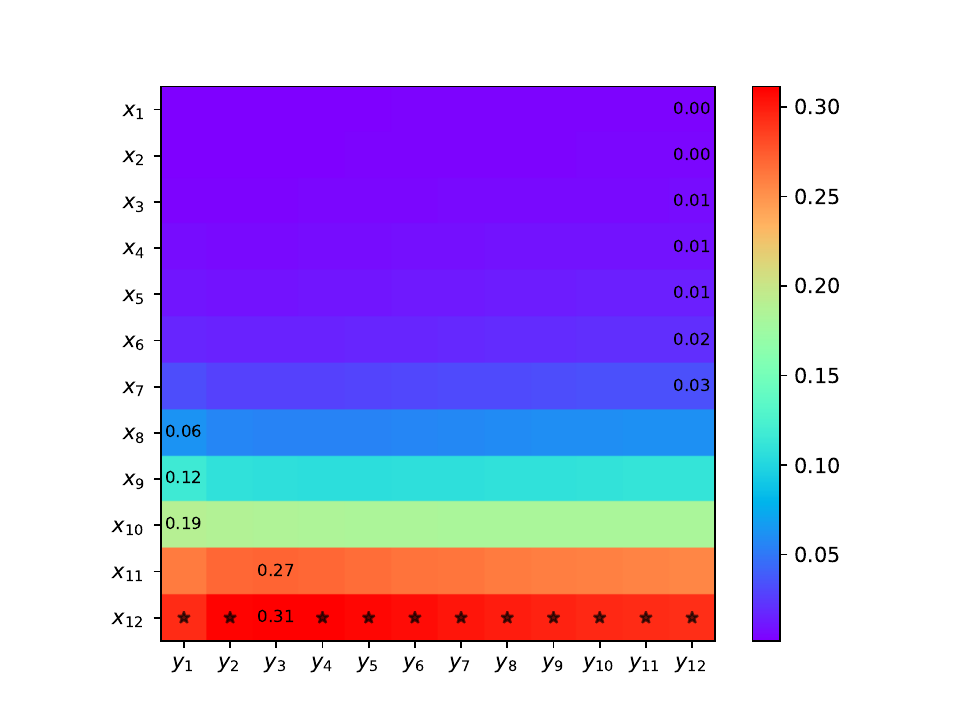}%
    \label{fig:attn_l_2}} \\
\subfloat[$k+1$]{%
    \includegraphics[trim=35 10 35 10,clip,width=.3\textwidth]{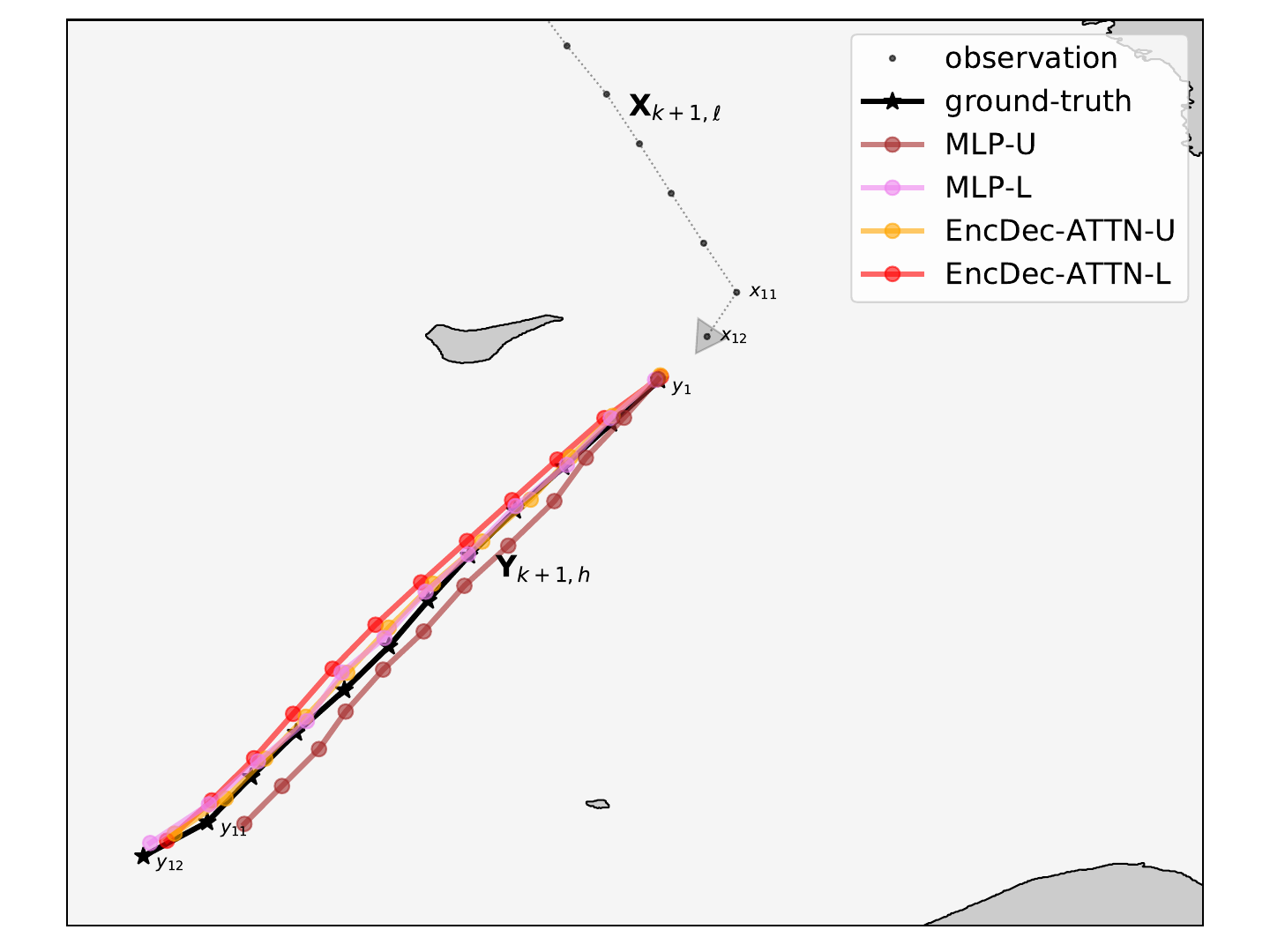}%
    \label{fig:prediction3}} \hfill
\subfloat[EncDec-ATTN-U ($k+1$)]{%
    \includegraphics[trim=50 20 50 35,clip,width=.3\textwidth]{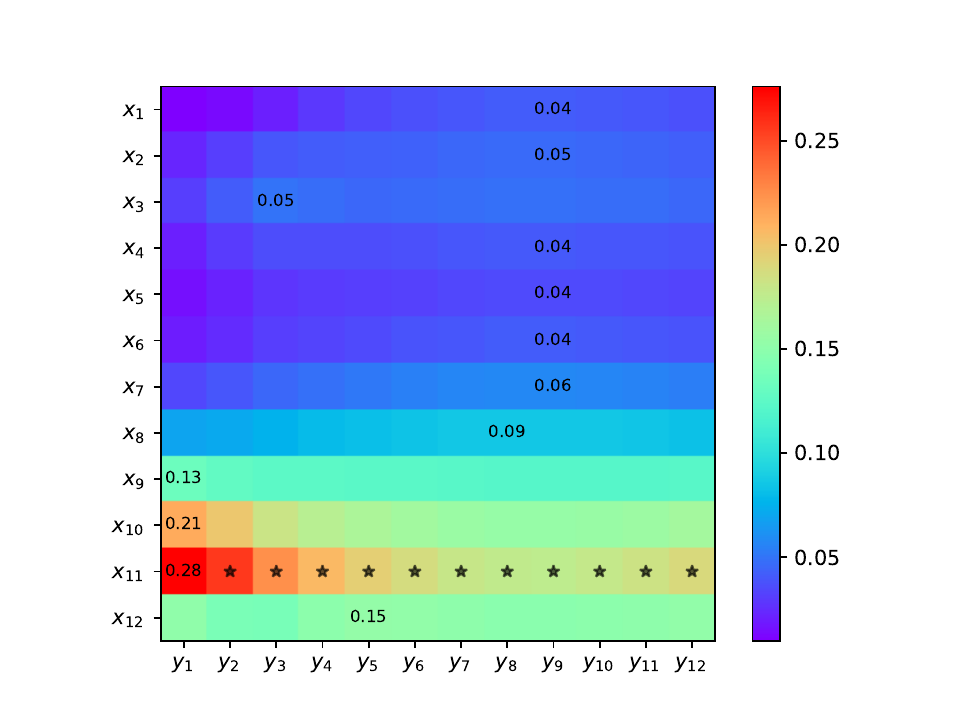}%
    \label{fig:attn_u_3}} \hfill
\subfloat[EncDec-ATTN-L ($k+1$)]{%
    \includegraphics[trim=50 20 50 35,clip,width=.3\textwidth]{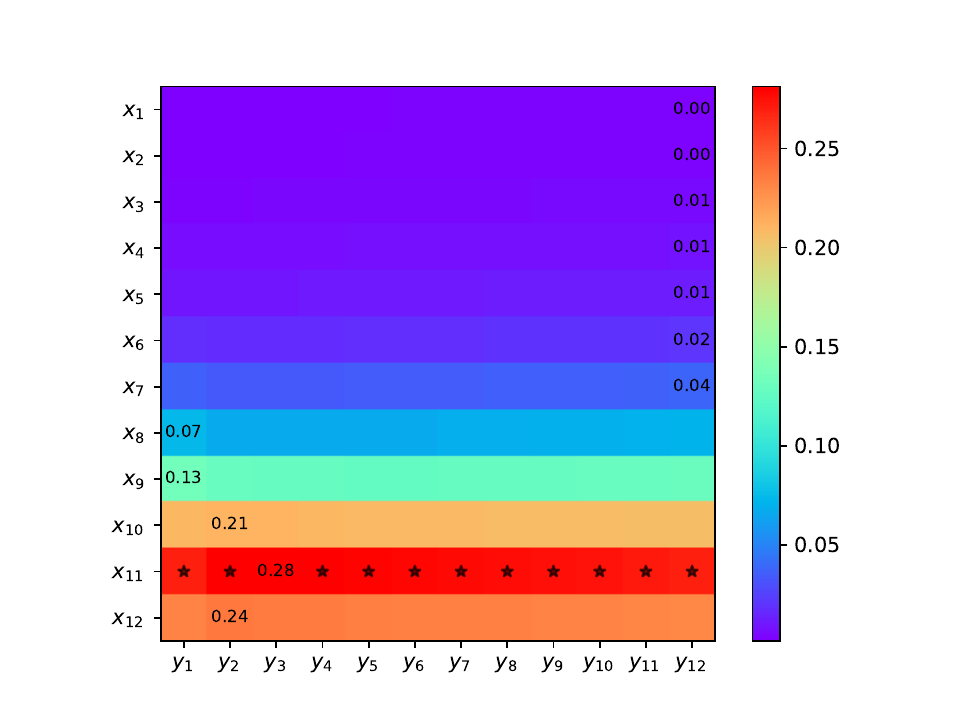}%
    \label{fig:attn_l_3}} \\
\subfloat[$k+2$]{%
    \includegraphics[trim=35 10 35 10,clip,width=.3\textwidth]{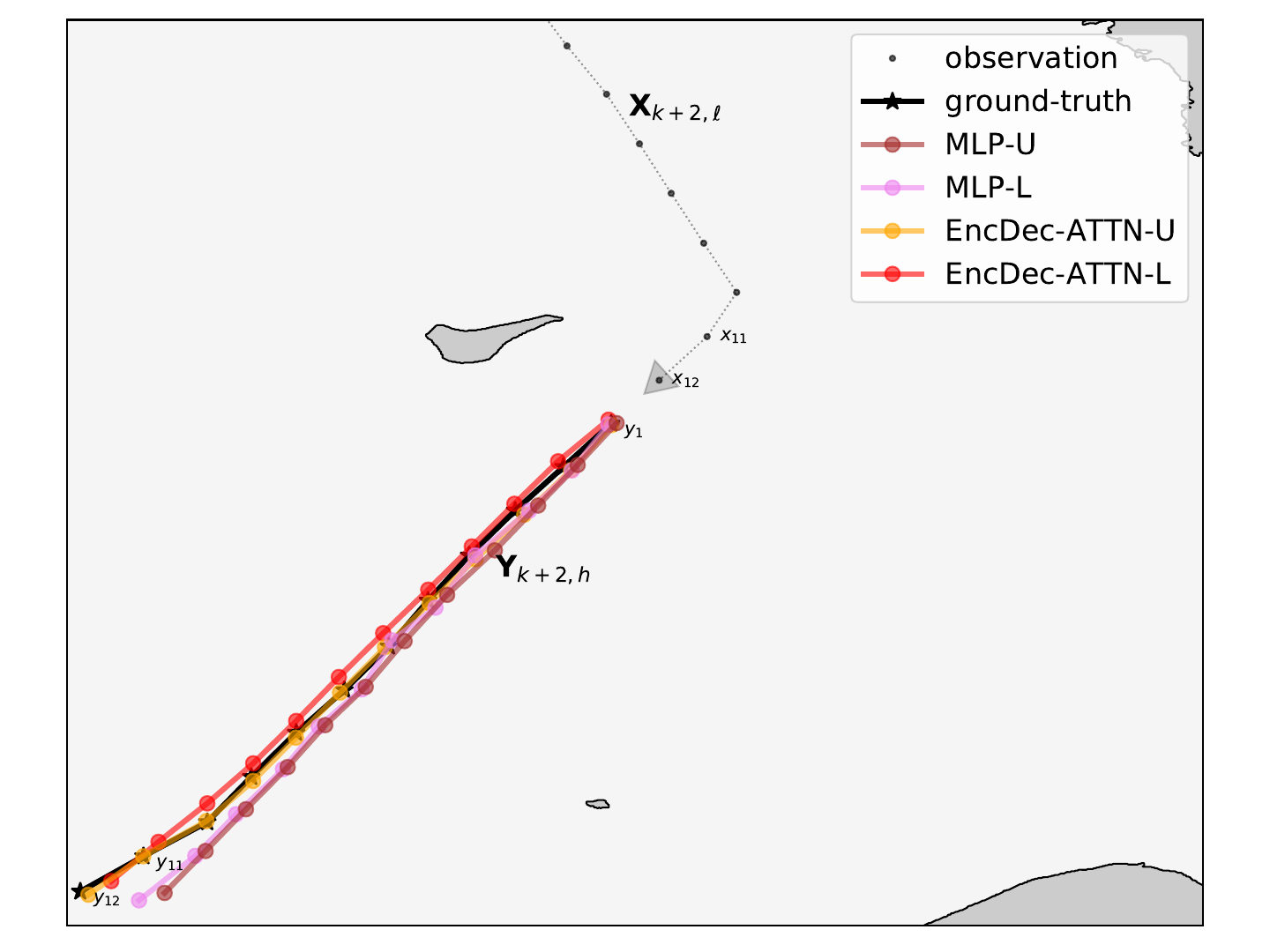}%
    \label{fig:prediction4}} \hfill
\subfloat[EncDec-ATTN-U ($k+2$)]{%
    \includegraphics[trim=50 20 50 35,clip,width=.3\textwidth]{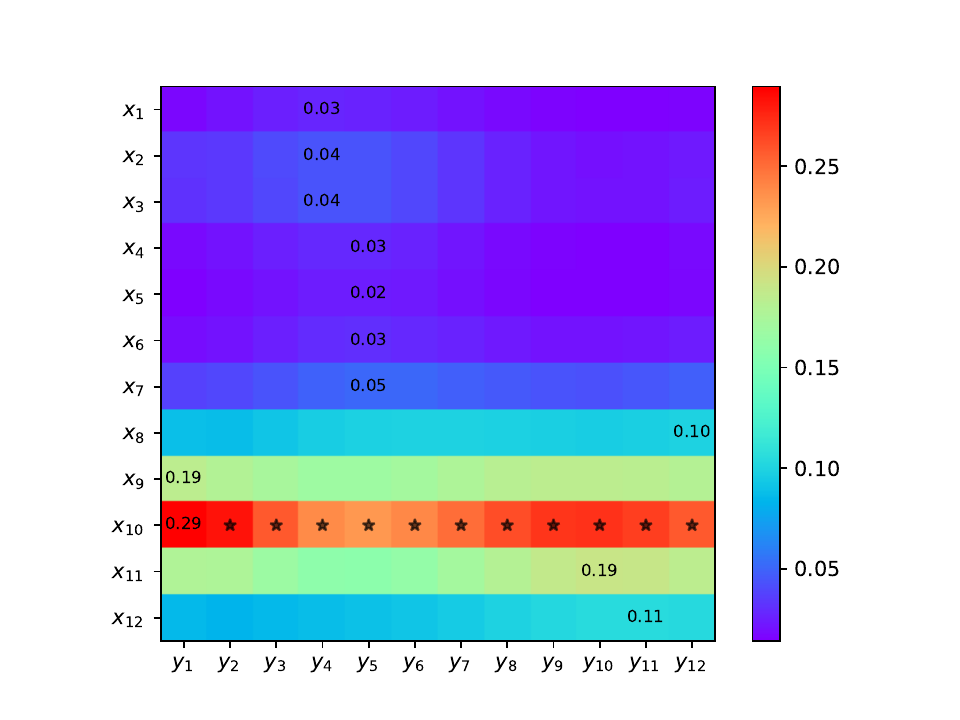}%
    \label{fig:attn_u_4}} \hfill
\subfloat[EncDec-ATTN-L ($k+2$)]{%
    \includegraphics[trim=50 20 50 35,clip,width=.3\textwidth]{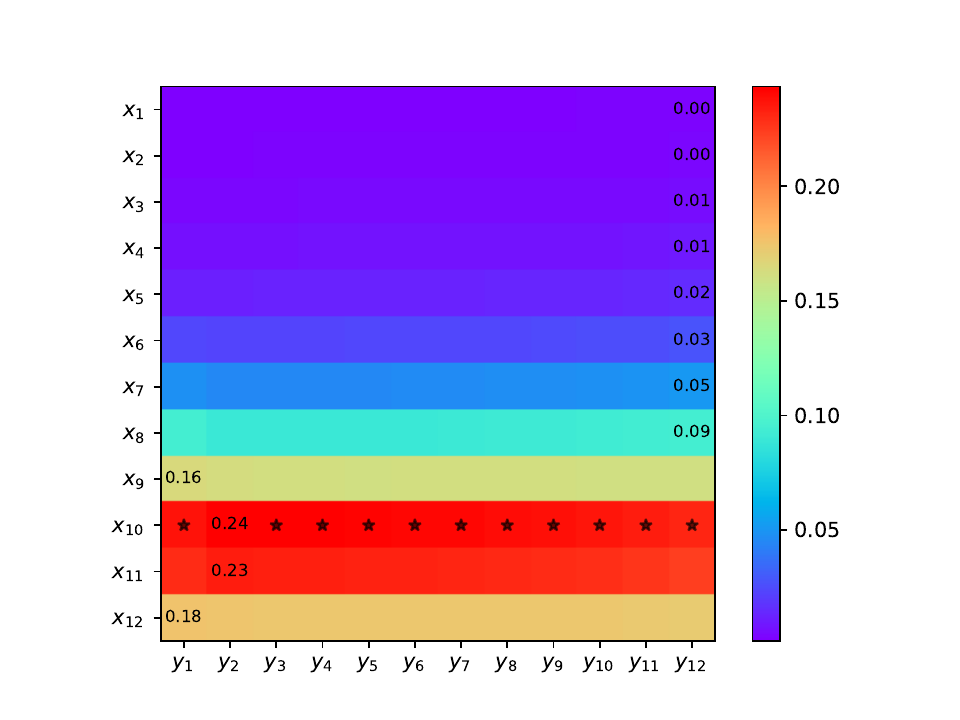}%
    \label{fig:attn_l_4}} 
\caption{Four different models map four input sequences $\mathbf{X}_{i, \ell}$ (from a journey which intersects the area $WP$ in Fig.~\ref{fig:compare_dist_b}) into their respective future trajectories $\hat{\mathbf{Y}}_{i, h}$ with $ i=k-1,k,k+1,k+2$. %
The temporal order of  input and  predicted sequences follows the index $k$ and increases row-wise from top to bottom. In this specific context, the labeled model outperforms the unlabeled one (for both the MLP and EncDec-ATTN architectures). \rev{The middle and right column illustrate the attention weight scores related to the predictions computed by the unlabeled (EncDec-ATTN-U) and labeled (EncDec-ATTN-L) recurrent models, respectively. In each panel, asterisks mark the maximum value on each column; instead, text labels indicate the maximum attention weight assigned to that position in relation to the output sequence. %
} %
\rev{Contains data from the Danish Maritime Authority that is used in accordance with the conditions for the use of Danish public data~\cite{DMA}.}} %
\label{fig:predictions}
\end{figure*}

\rev{Finally, in Fig.~\ref{fig:predictions} we show the results at four time steps (the predicted sequences in the left column, the attention weight scores from the unlabeled recurent model in the middle column, and the attention weight scores from the labeled recurrent model in the right column) obtained by four trained models on a specific trajectory, which intersects the area $WP$ (Fig.~\ref{fig:compare_dist_b}). The predictions are computed with the sliding window approach already discussed (see Section~\ref{se:segmentation}),  %
and the sequence of predictions follows the temporal index $k$ on the given trajectory.} We show that the labeled models \rev{(both MLP and EncDec-ATTN)} are able to forecast \rev{the vessel's position many time steps before the maneuvering point (see Figs.~\ref{fig:prediction1} and~\ref{fig:prediction2});}
conversely, %
the unlabeled models %
predict the correct direction only after the \rev{(branching)} critical \rev{position is observed} %
at time step $k$ \rev{(see Figs.~\ref{fig:prediction3} and~\ref{fig:prediction4})}. \rev{Intuitively, the labeled models can achieve better prediction performance because the additional information allows going from a higher modality prediction task to a lower modality one. In other words, the intention information enables labeled models to exclude well in advance paths that are incompatible with the actual vessel's destination.} %
\rev{Moreover, to quantify the error of the predictions shown in the left column of} Fig.~\ref{fig:predictions}, we measure the haversine distance $d_{H}(\hat{\mathbf{y}}_h^{k+1}, \mathbf{y}_h^{k+1})$ in nautical miles with a $3$-hour prediction horizon between the predicted position $\hat{\mathbf{y}}_h^{k+1}$ and the target position $\mathbf{y}_h^{k+1}$. %
\rev{Reading Table~\ref{tbl:new}, we also note that the EncDec-ATTN model in general achieves better prediction performance than the MLP. The only exception is at time $k+1$, when the labeled MLP is slightly better than the labeled recurrent model. In the following section, we explore how the attention mechanism can be used to interpret the neural model decisions.}

\begin{table}
\renewcommand{\arraystretch}{1.3}
\centering%
\caption{\rev{Evaluation of the prediction error in nautical miles at $3$ hours in the cases shown in the left column of Fig.~\ref{fig:predictions}.}}%
\label{tbl:new}
\begin{tabular}{r|cccc}
              & $k-1$ & $k$ & $k+1$ & $k+2$ \\ \hline
MLP-U         & \tablenum{30.32} & \tablenum{36.78}  & \tablenum{4.33}  &  \tablenum{3.14}   \\
EncDec-ATTN-U & \tablenum{30.31} & \tablenum{34.66}  & \tablenum{1.91}  &  \tablenum{0.37}   \\
MLP-L         & \tablenum{1.76} & \tablenum{1.68}  & \tablenum{0.92}  &  \tablenum{2.27}   \\
EncDec-ATTN-L & \tablenum{1.03} & \tablenum{1.22}  & \tablenum{1.36}  &  \tablenum{1.35}  
\end{tabular}
\end{table}

\subsection{\rev{Attention weight scores analysis}}

\rev{The attention mechanism also facilitates explanation of the internal working of a neural network by %
computing a vector of importance weights on the input sequence at each position in the predicted  sequence. In other words, since it computes a distribution over inputs, we can use the attention to provide an interpretation of model decisions~\cite{bahdanau2015, rush15, LeeSK17}.}

\rev{Inspired by prior work on attention interpretability~\cite{Serrano19, LeeSK17, rush15}, we can visualize the attention weights computed both by the unlabeled and labeled recurrent models on the same set of input/target sequences and attempt an interpretation of the importance the network gives to specific trajectory features.
Fig.~\ref{fig:predictions} is organized as follows; a single test trajectory is considered, from which we extracted four input/target sequences at consecutive time instants, one for each row in the figure, so that the time instant $k$ corresponds to the last input sequence before the waypoint. The left column shows the input, target and predicted sequences with the considered models; the middle and right columns show a visualization of the attention weight scores from the unlabeled and labeled models, respectively. 
More precisely, we visualize the attention weights $\alpha_{ij}$ in~\eqref{eq:attention}. Each matrix column in each plot indicates the weight distribution over the input sequence, so that it is possible to figure out which positions in the history pattern were considered more important when generating the predicted trajectory.
As time flows (from the top to the bottom row), the unlabeled model (middle column) is able to give more influence to the salient features of the input sequence while generating the output sequence. This is also apparent at visual inspection: the network gives more importance to the ending of the input sequence when computing the beginning of the output sequence (top two rows), but it is still able to recognize a maneuver in the input sequence and give more importance to the turning point; this is evident from the two bottom rows, where there is a brighter row that corresponds to the position of the turning point in the input sequence, which moves upwards as time flows and the position of the turning point shifts in the input sequence.
More specifically, at time instants $k-1$ and $k$, the unlabeled model gives more importance to the last position in the input sequence at the beginning of the predicted sequence, differently to what happens in the rest of the sequence, where the score is gradually reduced, with the score almost reducing to an average operation on the encoded features (Fig.~\ref{fig:attn_u_1} and~\ref{fig:attn_u_2}). In these cases, the predictions computed with the unlabeled models are also far away from the ground truth, as they follow the ``wrong'' path after the crossroad area.
A significantly different behavior is that of the labeled model, which gives more importance to the last five positions of the input sequence while generating the output sequence (Fig.~\ref{fig:attn_l_1} and~\ref{fig:attn_l_2}); in these cases, the network is able to pick the correct path after the waypoint, and the predicted trajectories are close to the ground truth.}

\rev{The behavior at time instants $k+1$ and $k+2$ is worth of note, as both the labeled and unlabeled models place greater emphasis on the positions in the input sequence that correspond to the area where the ship changed direction (waypoint). This is likely because even the unlabeled model is able to recognize the turning maneuver in the input sequence and thus put more emphasis on the input region where this pattern is clear. A major evidence of this behaviour is seen in the labeled model where the magnitude of weight score is very high in relation to that specific pattern in the input sequence (Fig.~\ref{fig:attn_l_3} and~\ref{fig:attn_l_4}).}

\section{\rev{Limitations and extensions}}
\label{sec:limitations}

\rev{The proposed encoder-decoder model presents some limitations which may be interesting to explore through future extensions. 
First, for reliable and accurate predictions, our method heavily relies upon historical data of paths belonging to the same motion pattern of the test trajectory,
i.e., a large and representative sample of data from the domain.
This means that the performance of our method may be sensitive to the training dataset, in particular, the number of ship trajectories available and domain representativeness.
The fact that the dataset may be an unrepresentative sample of data from the domain, and that
predictive model performance often improves with dataset size, are well-known issues. However, how to select the dataset is still a challenging open problem. This is usually addressed by performing a sensitivity analysis to quantify the relationship between dataset size or domain representativeness and model performance. 
It is also worth pointing out that, although the proposed encoder-decoder architecture achieves the best performance in the experiments, the MLP model can provide a good trade-off  between complexity and prediction performance, especially in the short term. However, this comes at the expense of diminished interpretability that the attention mechanism can provide, and the lowered ability to map an input sequence to an output sequence that is not necessarily of the same length.
As a matter of fact, compared to standard NN-based models, the encoder-decoder architecture allows training the RNNs to map a variable-length input sequence to another variable-length output sequence.
}

\rev{Another possible limitation concerns the use of high-level intention information in the labeled models.
As shown in the results, the intention information turns out to be very useful to solve the problem of multimodality of the prediction task when using unimodal forecasting, and exclude predictions that do not fulfill the constraint imposed by the intention.
However, such additional information is shown to be unhelpful when the multimodality involves a set of trajectories sharing the same intention.
This is apparent from Fig.~\ref{fig:prediction_distribution} where we can see how the labeled architecture on one hand can reduce the averaging effect between the two main branches compared to unlabeled models, but on the other hand, cannot completely address %
multimodality. Future extensions in this direction include the investigation of mixture density prediction techniques, which could definitely remove the averaging effect of %
patterns sharing the same intention information (i.e., destination). %
}

\section{Discussion and Conclusion}
\label{sec:conclusion}

In this paper, we presented a recurrent encoder-decoder architecture to address the problem of %
trajectory prediction in the 
presence of complex mobility patterns, testing the approach in a maritime domain case study.
An extensive experimental campaign with real AIS data showed that the proposed architecture is able to learn space-time dependencies from historical ship mobility data and successfully predict future vessel trajectories.
Experimental results show the superiority of attention-based models over simple aggregation functions, %
and 
that a significant performance improvement %
can be achieved with the use of information on the vessel's destination.
Future work will focus on  
extending the proposed encoder-decoder architecture to directly address the multi-modal nature of the vessel prediction task, as well as on modeling the prediction uncertainty. %

\ifCLASSOPTIONcaptionsoff
  \newpage
\fi

\bibliographystyle{IEEEtran}
\bibliography{IEEEabrv,refs_clean_final}

% Generated by IEEEtran.bst, version: 1.14 (2015/08/26)
\begin{thebibliography}{10}
\providecommand{\url}[1]{#1}
\csname url@samestyle\endcsname
\providecommand{\newblock}{\relax}
\providecommand{\bibinfo}[2]{#2}
\providecommand{\BIBentrySTDinterwordspacing}{\spaceskip=0pt\relax}
\providecommand{\BIBentryALTinterwordstretchfactor}{4}
\providecommand{\BIBentryALTinterwordspacing}{\spaceskip=\fontdimen2\font plus
\BIBentryALTinterwordstretchfactor\fontdimen3\font minus
  \fontdimen4\font\relax}
\providecommand{\BIBforeignlanguage}[2]{{%
\expandafter\ifx\csname l@#1\endcsname\relax
\typeout{** WARNING: IEEEtran.bst: No hyphenation pattern has been}%
\typeout{** loaded for the language `#1'. Using the pattern for}%
\typeout{** the default language instead.}%
\else
\language=\csname l@#1\endcsname
\fi
#2}}
\providecommand{\BIBdecl}{\relax}
\BIBdecl

\bibitem{DMA}
``Data from the {D}anish {AIS} system,''
  \url{https://www.dma.dk/SikkerhedTilSoes/Sejladsinformation/AIS/Sider/default.aspx}.

\bibitem{Mozaffari2020}
S.~{Mozaffari}, O.~Y. {Al-Jarrah}, M.~{Dianati}, P.~{Jennings}, and
  A.~{Mouzakitis}, ``Deep learning-based vehicle behavior prediction for
  autonomous driving applications: A review,'' \emph{{IEEE} Trans. Intell.
  Transp. Syst.}, 2020.

\bibitem{Rong2003}
X.~{Rong Li} and V.~P. {Jilkov}, ``Survey of maneuvering target tracking. {Part
  I}. {Dynamic} models,'' \emph{{IEEE} Trans. Aerosp. Electron. Syst.},
  vol.~39, no.~4, pp. 1333--1364, 2003.

\bibitem{Millefiori2016}
L.~M. {Millefiori}, P.~{Braca}, K.~{Bryan}, and P.~{Willett}, ``Modeling vessel
  kinematics using a stochastic mean-reverting process for long-term
  prediction,'' \emph{{IEEE} Trans. Aerosp. Electron. Syst.}, vol.~52, no.~5,
  pp. 2313--2330, 2016.

\bibitem{Millefiori2015}
L.~M. {Millefiori}, G.~{Pallotta}, P.~{Braca}, S.~{Horn}, and K.~{Bryan},
  ``Validation of the {O}rnstein-{U}hlenbeck route propagation model in the
  {M}editerranean {S}ea,'' in \emph{IEEE/MTS OCEANS}, 2015.

\bibitem{Uney2019}
M.~{Üney}, L.~M. {Millefiori}, and P.~{Braca}, ``Data driven vessel trajectory
  forecasting using stochastic generative models,'' in \emph{IEEE International
  Conference on Acoustics, Speech and Signal Processing (ICASSP)}, 2019, pp.
  8459--8463.

\bibitem{icassp19}
N.~{Forti}, L.~M. {Millefiori}, P.~{Braca}, and P.~{Willett}, ``Anomaly
  detection and tracking based on mean--reverting processes with unknown
  parameters,'' in \emph{IEEE International Conference on Acoustics, Speech and
  Signal Processing (ICASSP)}, 2019, pp. 8449--8453.

\bibitem{Millefiori2016-2}
L.~M. Millefiori, P.~Braca, and P.~Willett, ``Consistent estimation of randomly
  sampled {O}rnstein-{U}hlenbeck process long-run mean for long-term target
  state prediction,'' \emph{{IEEE} Signal Process. Lett.}, vol.~23, no.~11, pp.
  1562--1566, 2016.

\bibitem{Coscia2018}
P.~{Coscia}, P.~{Braca}, L.~M. {Millefiori}, F.~A.~N. {Palmieri}, and
  P.~{Willett}, ``Multiple {Ornstein–Uhlenbeck} processes for maritime
  traffic graph representation,'' \emph{{IEEE} Trans. Aerosp. Electron. Syst.},
  vol.~54, no.~5, pp. 2158--2170, 2018.

\bibitem{oceans2019}
N.~Forti, L.~M. Millefiori, and P.~Braca, ``Unsupervised extraction of maritime
  patterns of life from {A}utomatic {I}dentification {S}ystem data,'' in
  \emph{IEEE/MTS OCEANS}, 2019.

\bibitem{Ristic08}
B.~{Ristic}, B.~{La Scala}, M.~{Morelande}, and N.~{Gordon}, ``Statistical
  analysis of motion patterns in {AIS} data: Anomaly detection and motion
  prediction,'' in \emph{International Conference on Information Fusion
  (FUSION)}, 2008.

\bibitem{Hexeberg17}
S.~{Hexeberg}, A.~L. {Flåten}, B.~H. {Eriksen}, and E.~F. {Brekke},
  ``{AIS}-based vessel trajectory prediction,'' in \emph{International
  Conference on Information Fusion (FUSION)}, 2017.

\bibitem{Mazzarella15}
F.~{Mazzarella}, V.~F. {Arguedas}, and M.~{Vespe}, ``Knowledge-based vessel
  position prediction using historical {AIS} data,'' in \emph{Sensor Data
  Fusion: Trends, Solutions, Applications}, 2015.

\bibitem{Rong2019}
H.~Rong, A.~Teixeira, and C.~{Guedes Soares}, ``Ship trajectory uncertainty
  prediction based on a {Gaussian Process} model,'' \emph{Ocean Engineering},
  vol. 182, pp. 499--511, 2019.

\bibitem{Zissis2017}
A.~Valsamis, K.~Tserpes, D.~Zissis, D.~Anagnostopoulos, and T.~Varvarigou,
  ``Employing traditional machine learning algorithms for big data streams
  analysis: The case of object trajectory prediction,'' \emph{Journal of
  Systems and Software}, vol. 127, pp. 249--257, 2017.

\bibitem{survey2018}
E.~{Tu}, G.~{Zhang}, L.~{Rachmawati}, E.~{Rajabally}, and G.~{Huang},
  ``Exploiting {AIS} data for intelligent maritime navigation: A comprehensive
  survey from data to methodology,'' \emph{{IEEE} Trans. Intell. Transp.
  Syst.}, vol.~19, no.~5, pp. 1559--1582, 2018.

\bibitem{survey2020}
Z.~{Xiao}, X.~{Fu}, L.~{Zhang}, and R.~S.~M. {Goh}, ``Traffic pattern mining
  and forecasting technologies in maritime traffic service networks: A
  comprehensive survey,'' \emph{{IEEE} Trans. Intell. Transp. Syst.}, vol.~21,
  no.~5, pp. 1796--1825, 2020.

\bibitem{Nguyen2018}
D.~Nguyen, R.~Vadaine, G.~Hajduch, R.~Garello, and R.~Fablet, ``A multi-task
  deep learning architecture for maritime surveillance using {AIS} data
  streams,'' in \emph{{IEEE} International Conference on Data Science and
  Advanced Analytics (DSAA)}, 2018, pp. 331--340.

\bibitem{Gao2018}
M.~Gao, G.~Shi, and S.~Li, ``Online prediction of ship behavior with automatic
  identification system sensor data using bidirectional long short-term memory
  recurrent neural network,'' \emph{Sensors}, vol.~18, no.~12, 2018.

\bibitem{Yu2020}
J.~Y. {Yu}, M.~O. {Sghaier}, and Z.~{Grabowiecka}, ``Deep learning approaches
  for {AIS} data association in the context of maritime domain awareness,'' in
  \emph{International Conference on Information Fusion (FUSION)}, 2020.

\bibitem{Murray2020}
B.~Murray and L.~P. Perera, ``A dual linear autoencoder approach for vessel
  trajectory prediction using historical {AIS} data,'' \emph{Ocean
  Engineering}, vol. 209, p. 107478, 2020.

\bibitem{Zhou2020}
X.~Zhou, Z.~Liu, F.~Wang, Y.~Xie, and X.~Zhang, ``Using deep learning to
  forecast maritime vessel flows,'' \emph{Sensors}, vol.~20, no.~6, 2020.

\bibitem{Nguyen2018b}
D.-D. Nguyen, C.~L. Van, and M.~I. Ali, ``Vessel trajectory prediction using
  sequence-to-sequence models over spatial grid,'' in \emph{ACM International
  Conference on Distributed and Event-based Systems (DEBS)}, 2018, pp.
  258--261.

\bibitem{Forti2020}
N.~{Forti}, L.~M. {Millefiori}, P.~{Braca}, and P.~{Willett}, ``Prediction of
  vessel trajectories from {AIS} data via sequence-to-sequence recurrent neural
  networks,'' in \emph{IEEE International Conference on Acoustics, Speech and
  Signal Processing (ICASSP)}, 2020, pp. 8936--8940.

\bibitem{Jung2020}
S.~Jung, I.~Schlangen, and A.~Charlish, ``A mnemonic kalman filter for
  non-linear systems with extensive temporal dependencies,'' \emph{{IEEE}
  Signal Process. Lett.}, vol.~27, pp. 1005--1009, 2020.

\bibitem{Cho2014}
K.~Cho, B.~Van~Merri{\"e}nboer, C.~Gulcehre, D.~Bahdanau, F.~Bougares,
  H.~Schwenk, and Y.~Bengio, ``Learning phrase representations using {RNN}
  encoder-decoder for statistical machine translation,'' in \emph{Conference on
  Empirical Methods in Natural Language Processing (EMNLP)}, 2014, pp.
  1724--1734.

\bibitem{Sutskever2014}
I.~Sutskever, O.~Vinyals, and Q.~V. Le, ``Sequence to sequence learning with
  neural networks,'' in \emph{International Conference on Neural Information
  Processing Systems (NIPS)}, 2014, pp. 3104--3112.

\bibitem{Chiu18}
C.~{Chiu} \emph{et~al.}, ``State-of-the-art speech recognition with
  sequence-to-sequence models,'' in \emph{IEEE International Conference on
  Acoustics, Speech and Signal Processing (ICASSP)}, 2018, pp. 4774--4778.

\bibitem{Hochreiter1997}
S.~Hochreiter and J.~Schmidhuber, ``Long short-term memory,'' \emph{Neural
  computation}, vol.~9, no.~8, pp. 1735--1780, 1997.

\bibitem{EMODNET_detail}
L.~Falco, A.~Pititto, W.~Adnams, N.~Earwaker, and H.~Greidanus, ``{EU} vessel
  density map -- detailed method,''
  \url{https://www.emodnet-humanactivities.eu/documents/Vessel\%20density\%20maps_method_v1.5.pdf},
  EMODnet, Tech. Rep., 2019.

\bibitem{EMODNET_HA}
EMODnet, ``Human activities portal,''
  \url{https://www.emodnet-humanactivities.eu}.

\bibitem{Hancock2020}
J.~T. Hancock and T.~M. Khoshgoftaar, ``Survey on categorical data for neural
  networks,'' \emph{J. Big Data}, vol.~7, no.~1, p.~28, 2020.

\bibitem{piech2015deep}
C.~Piech, J.~Bassen, J.~Huang, S.~Ganguli, M.~Sahami, L.~J. Guibas, and
  J.~Sohl-Dickstein, ``Deep knowledge tracing,'' in \emph{Advances in neural
  information processing systems}, 2015, pp. 505--513.

\bibitem{Zaroug20}
A.~Zaroug, D.~T.~H. Lai, K.~Mudie, and R.~Begg, ``Lower limb kinematics
  trajectory prediction using long short-term memory neural networks,''
  \emph{Frontiers in Bioengineering and Biotechnology}, vol.~8, p. 362, 2020.

\bibitem{Bishop96}
C.~M. Bishop, \emph{Neural Networks for Pattern Recognition}.\hskip 1em plus
  0.5em minus 0.4em\relax Oxford University Press, Inc., 1996.

\bibitem{Graves2013}
A.~{Graves}, A.~{Mohamed}, and G.~{Hinton}, ``Speech recognition with deep
  recurrent neural networks,'' in \emph{IEEE International Conference on
  Acoustics, Speech and Signal Processing (ICASSP)}, 2013, pp. 6645--6649.

\bibitem{bahdanau2015}
D.~Bahdanau, K.~Cho, and Y.~Bengio, ``Neural machine translation by jointly
  learning to align and translate,'' in \emph{International Conference on
  Learning Representations (ICLR)}, 2015.

\bibitem{Xu2015}
K.~Xu, J.~Ba, R.~Kiros, K.~Cho, A.~C. Courville, R.~Salakhutdinov, R.~S. Zemel,
  and Y.~Bengio, ``Show, attend and tell: Neural image caption generation with
  visual attention,'' in \emph{International Conference on Machine Learning
  (ICML)}, 2015, pp. 2048--2057.

\bibitem{Bengio1994}
Y.~Bengio, P.~Y. Simard, and P.~Frasconi, ``Learning long-term dependencies
  with gradient descent is difficult,'' \emph{{IEEE} Trans. Neural Netw.},
  vol.~5, no.~2, pp. 157--166, 1994.

\bibitem{Felix2000}
F.~A. Gers, J.~Schmidhuber, and F.~Cummins, ``Learning to forget: Continual
  prediction with {LSTM},'' \emph{Neural Computation}, vol.~12, no.~10, pp.
  2451--2471, 2000.

\bibitem{Schuster97}
M.~Schuster and K.~K. Paliwal, ``Bidirectional recurrent neural networks,''
  \emph{{IEEE} Trans. Signal Process.}, vol.~45, no.~11, pp. 2673--2681, 1997.

\bibitem{Graves2005}
A.~Graves and J.~Schmidhuber, ``Framewise phoneme classification with
  bidirectional {LSTM} and other neural network architectures,'' \emph{Neural
  Networks}, vol.~18, no. 5-6, pp. 602--610, 2005.

\bibitem{Boureau2010}
Y.-L. Boureau, J.~Ponce, and Y.~LeCun, ``A theoretical analysis of feature
  pooling in visual recognition,'' in \emph{International Conference on Machine
  Learning (ICML)}, 2010, pp. 111–--118.

\bibitem{Martina2020}
S.~{Martina}, L.~{Ventura}, and P.~{Frasconi}, ``Classification of cancer
  pathology reports: A large-scale comparative study,'' \emph{{IEEE} J. Biomed.
  Health Inform.}, vol.~24, no.~11, pp. 3085--3094, 2020.

\bibitem{Hermann2015}
K.~M. Hermann, T.~Kocisk{\'{y}}, E.~Grefenstette, L.~Espeholt, W.~Kay,
  M.~Suleyman, and P.~Blunsom, ``Teaching machines to read and comprehend,'' in
  \emph{International Conference on Neural Information Processing Systems
  (NIPS)}, 2015, pp. 1693--1701.

\bibitem{AIS_ITU}
\emph{Technical characteristics for an automatic identification system using
  time-division multiple access in the {VHF} maritime mobile band}, {ITU}
  Recommendation {M.1371}, Rev.~5, Feb. 2014.

\bibitem{Gareth2014}
G.~James, D.~Witten, T.~Hastie, and R.~Tibshirani, \emph{An Introduction to
  Statistical Learning: With Applications in R}.\hskip 1em plus 0.5em minus
  0.4em\relax Springer New York, 2014.

\bibitem{Goodfellow2016}
I.~Goodfellow, Y.~Bengio, and A.~Courville, \emph{Deep Learning}.\hskip 1em
  plus 0.5em minus 0.4em\relax MIT Press, 2016.

\bibitem{Glorot11}
X.~Glorot, A.~Bordes, and Y.~Bengio, ``Deep sparse rectifier neural networks,''
  in \emph{International Conference on Artificial Intelligence and Statistics
  (AISTATS)}, 2011, pp. 315--323.

\bibitem{Kingma2014}
D.~P. Kingma and J.~Ba, ``Adam: {A} method for stochastic optimization,'' in
  \emph{International Conference on Learning Representations (ICLR)}, 2015.

\bibitem{Goodfellow16}
I.~Goodfellow, Y.~Bengio, and A.~Courville, \emph{Deep Learning}.\hskip 1em
  plus 0.5em minus 0.4em\relax MIT Press, 2016,
  \url{http://www.deeplearningbook.org}.

\bibitem{Kaiming2015}
K.~He, X.~Zhang, S.~Ren, and J.~Sun, ``Delving deep into rectifiers: Surpassing
  human-level performance on imagenet classification,'' in \emph{IEEE
  International Conference on Computer Vision (ICCV)}, 2015, pp. 1026–--1034.

\bibitem{Henaff2016}
M.~Henaff, A.~Szlam, and Y.~LeCun, ``Recurrent orthogonal networks and
  long-memory tasks,'' in \emph{International Conference on Machine Learning
  (ICML)}, 2016, pp. 2034--2042.

\bibitem{Vorontsov2017}
E.~Vorontsov, C.~Trabelsi, S.~Kadoury, and C.~Pal, ``On orthogonality and
  learning recurrent networks with long term dependencies,'' in
  \emph{International Conference on Machine Learning (ICML)}, 2017, pp.
  3570--3578.

\bibitem{Glorot2010}
X.~Glorot and Y.~Bengio, ``Understanding the difficulty of training deep
  feedforward neural networks,'' in \emph{International Conference on
  Artificial Intelligence and Statistics (AISTATS)}, 2010, pp. 249--256.

\bibitem{Rafal2015}
R.~Jozefowicz, W.~Zaremba, and I.~Sutskever, ``An empirical exploration of
  recurrent network architectures,'' in \emph{International Conference on
  Machine Learning (ICML)}, 2015, pp. 2342–--2350.

\bibitem{Pellegrini2009}
S.~Pellegrini, A.~Ess, K.~Schindler, and L.~V. Gool, ``You'll never walk alone:
  Modeling social behavior for multi-target tracking,'' in \emph{{IEEE}
  International Conference on Computer Vision (ICCV)}, 2009, pp. 261--268.

\bibitem{Hastie2009}
T.~Hastie, R.~Tibshirani, and J.~Friedman, \emph{The Elements of Statistical
  Learning}.\hskip 1em plus 0.5em minus 0.4em\relax Springer New York, 2009.

\bibitem{rush15}
A.~M. Rush, S.~Chopra, and J.~Weston, ``A neural attention model for
  abstractive sentence summarization,'' in \emph{Conference on Empirical
  Methods in Natural Language Processing (EMNLP)}, Sep. 2015, pp. 379--389.

\bibitem{LeeSK17}
J.~Lee, J.~Shin, and J.~Kim, ``Interactive visualization and manipulation of
  attention-based neural machine translation,'' in \emph{Conference on
  Empirical Methods in Natural Language Processing (EMNLP)}, 2017, pp.
  121--126.

\bibitem{Serrano19}
S.~Serrano and N.~A. Smith, ``Is attention interpretable?'' in \emph{Conference
  of the Association for Computational Linguistics (ACL)}, 2019, pp.
  2931--2951.

\end{thebibliography}

\end{document}